\documentclass[11pt,a4paper,gdm]{google}
\usepackage[authoryear,sort&compress,round]{natbib}
\usepackage{amsmath,amsfonts,amssymb}
\usepackage{graphicx}
\usepackage{booktabs}
\usepackage{multirow}
\usepackage{xspace}
\usepackage[most]{tcolorbox}

\hypersetup{
  pdftitle={The Flip Test: Video World Models Fail to Learn the World State},
  pdfauthor={Mi Luo, Lijun Yu, Adams Wei Yu, Ang Cao, Baochen Sun, Ming-Hsuan Yang, Nan Hua, Yuan Hao, Alex Dimakis, Ya Xu, Ed H. Chi, Xuezhi Wang, Yinxiao Liu}
}

\usepackage[capitalize]{cleveref}
\crefname{section}{Sec.}{Secs.}
\Crefname{section}{Section}{Sections}
\Crefname{table}{Table}{Tables}
\crefname{table}{Tab.}{Tabs.}

\newcommand{\nolastrunt}{\parfillskip=0pt plus 0.7\linewidth\relax}
\AtBeginDocument{\nolastrunt}
\DeclareCaptionJustification{nolastrunt}{\nolastrunt}
\setlist[itemize]{noitemsep,leftmargin=*,topsep=0em}
\setlist[enumerate]{noitemsep,leftmargin=*,topsep=0em}

\makeatletter
\DeclareRobustCommand\onedot{\futurelet\@let@token\@onedot}
\def\@onedot{\ifx\@let@token.\else.\null\fi\xspace}
\def\eg{\emph{e.g}\onedot}

\makeatother

\definecolor{ftblue}{HTML}{4A6FAE}
\definecolor{ftteal}{HTML}{2E8F8E}
\definecolor{ftbar}{HTML}{80A1D4}
\definecolor{ftbg}{HTML}{F4F2F8}
\definecolor{ftpanel}{RGB}{47,51,80}
\newcommand{\fttable}{\small\setlength{\tabcolsep}{5pt}\setlength{\aboverulesep}{0pt}\setlength{\belowrulesep}{0pt}\renewcommand{\arraystretch}{1.28}}
\newcommand{\fthead}{\rowcolor{ftblue!13}}
\newcommand{\ftzebra}{\rowcolor{ftbg}}

\newtcolorbox{keyidea}{enhanced, frame hidden, colback=ftbg, arc=2pt, boxrule=0pt,
  borderline west={2.2pt}{0pt}{ftbar}, left=7pt, right=6pt, top=4pt, bottom=4pt,
  before skip=7pt, after skip=7pt, fontupper=\itshape\small, before upper={\nolastrunt}}
\DeclareRobustCommand{\levelimg}[1]{\raisebox{-0.25em}{\includegraphics[height=1.15em]{fig/level#1.pdf}}\,}
\newcommand{\levelicon}[1]{\texorpdfstring{\levelimg{#1}}{}}
\newcommand{\finding}[1]{{\upshape\bfseries\textcolor{ftblue}{F#1:}}~}

\newcommand{\promptbox}[3]{\fcolorbox{black!35}{black!3}{\begin{minipage}[t]{\dimexpr#1\linewidth-2\fboxsep-2\fboxrule\relax}\raggedright\scriptsize\textbf{\sffamily #2}\par\vspace{3pt}{\ttfamily #3}\par\end{minipage}}}

\title{\texorpdfstring{\raisebox{-0.28\height}{\includegraphics[height=1.6em]{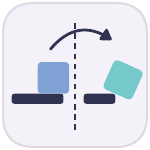}}\hspace{0.35em}}{}The Flip Test: \\ Video World Models Fail to Learn the World State}

\uselogo{}
\makeatletter
\renewcommand{\titlefont}{\color{black}\normalfont\bfseries\fontsize{19}{22}\selectfont}
\makeatother

\author[1,2,*]{Mi~Luo}
\author[2,$\dagger$]{Lijun~Yu}
\author[2]{Adams~Wei~Yu}
\author[2]{Ang~Cao}
\author[2]{Baochen~Sun}
\author[2]{Ming-Hsuan~Yang}
\author[2]{Nan~Hua}
\author[2]{Yuan~Hao}
\author[3]{Alex~Dimakis}
\author[2]{Ya~Xu}
\author[2]{Ed~H.~Chi}
\author[2,$\dagger$]{Xuezhi~Wang}
\author[2,$\dagger$]{Yinxiao~Liu}
\affil[1]{UT Austin}
\affil[2]{Google DeepMind}
\affil[3]{UC Berkeley}
\makeatletter
\def\AB@affilsep{\protect\\\protect\Affilfont}
\makeatother
\affil[*]{Work done during an internship at Google DeepMind}
\affil[$\dagger$]{Project co-lead}

\begin{abstract}
Video world models promise to predict the future of a scene from its present.
Do current models keep this promise?
We probe this question counterfactually.
When a small change to the present flips the future, such as a box resting a few millimeters before or past a table edge, a model that tracks the state of the world must flip its prediction too.
We build \textbf{FlipTest}, a three-level hierarchy of increasingly strict tests over 1{,}680 initial frames from 10 simulated and 15 real-world tasks.
At the base, \emph{visual minimal pairs}, two nearly identical frames whose true futures differ, test whether predictions change across the boundary; matched controls go further and test whether they change only where the future does; and dense sweeps, the strictest level, locate exactly where they change.
All 11 video world models we evaluate fall short.
The best averages 18.3\% pair accuracy over real and simulated scenes, below the 25\% of random guessing.
Models flip as often on steps that leave the future unchanged as on steps that change it, and their response stays flat across the sweeps.
On the same frames, a vision-language model is selective and traces a near S-shaped curve near the boundary, and a coding agent that measures the image comes close to the true boundary.
This implies that the deciding information is in the frame, and video models just miss it.
We therefore urge the community to rethink world modeling as pure video prediction.
Next-frame prediction must do more than draw plausible pixels; it must also identify the causal states that control future dynamics.
\par\vspace{0.7em}\noindent{\parfillskip=0pt plus 1fil\relax Project page: \href{https://flip-test.github.io/}{https://flip-test.github.io/}\par}
\end{abstract}

\begin{document}

\maketitle
\suppressfloats[t]
\makeatletter
\gdef\@title{The Flip Test: Video World Models Fail to Learn the World State}
\makeatother

\begin{figure}[t]
    {\includegraphics[width=0.49395\linewidth, trim=2.88 113.24 66 0, clip]{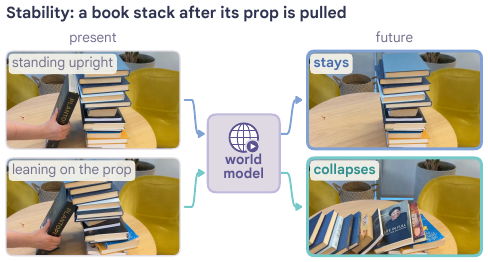}\par}
    \nointerlineskip
    \centering
    \includegraphics[width=0.7\linewidth, trim=0 0 0 12.4, clip]{fig/fig1-teaser.pdf}
    \caption{\textbf{A small change to the initial frame flips the correct future. }
Two nearly identical initial frames sit on opposite sides of a physical boundary, so the correct outcome differs.
A model passes only if its two predictions also differ.}
    \label{fig:fig1-teaser}
\end{figure}

\section{Introduction}
\label{sec:intro}

\emph{The world is causal.}
Whether a cup near a table edge falls, a ball passes through a narrow gap, or a weighted beam tips is already determined in the present by a few millimeters of position.
Humans learn early to read these futures directly from the present state.
Infants just a few months old look longer when a precariously perched box fails to fall~\citep{baillargeon1992support, baillargeon2004infants}; they expect objects to behave as solid, continuous bodies~\citep{spelke1992origins} and actively explore events that violate these expectations~\citep{stahl2015observing}.
Craik proposed that the mind achieves this by building a small-scale \emph{world model} of external reality to mentally test alternatives before acting~\citep{craik1967nature}.
Later work grounded these models in motor control~\citep{wolpert1995internal} and intuitive physics, where dedicated neural circuits~\citep{fischer2016functional} run approximate mental simulations of scenes~\citep{hegarty2004mechanical, battaglia2013simulation, ullman2017mind, kubricht2017intuitive}.
Crucially, an internal \emph{world model} is useful precisely because its forecasts are causally tied to the present: a critical shift in initial conditions yields a fundamentally different future.

Modern AI aspires to endow machines with this same predictive power.
Embodied agents learn internal dynamics to plan inside latent worlds~\citep{ha2018world, hafner2025mastering}, while large-scale video world models aim to simulate the future directly in pixels~\citep{sora, cosmos1, cosmos3}.
Trained to extrapolate a scene from its initial frame, today's video generators produce photorealistic rollouts and are increasingly championed as general-purpose simulators for robotics and embodied AI~\citep{nvidia2026gr00t2, waymo2026worldmodel}.
Yet, visual realism alone does not guarantee physical understanding.
Rendering plausible-looking motion falls short of the core promise of a true world model: \emph{predicting the specific future that the present implies}.

How can we rigorously test whether video world models fulfill this promise?
Existing benchmarks evaluate generated videos in isolation, scoring each rollout for general physical plausibility~\citep{bansal2025videophy, videophy2, physicsiq, worldmodelbench}.
However, plausibility is notoriously ill-defined and requires bespoke, event-specific metrics that are difficult to scale.
More importantly, single-video evaluation has a critical blind spot: a model can hallucinate the exact same convincing rollout for two scenes whose true futures are opposite, and an isolated plausibility score will never catch the contradiction.
Our key insight is to instead probe world models \emph{counterfactually}---by \emph{perturbing the present just enough to flip the future}.
By comparing predictions across such tipping points, we reduce evaluation to a clean, binary outcome label that generalizes across domains with a sharp physical boundary.

Building on this counterfactual principle, we introduce \textbf{FlipTest}, a three-level pyramid of increasingly stringent diagnostics (\cref{fig:fig1-teaser}).
FlipTest comprises 1{,}680 initial frames spanning 10 simulated and 15 real-world tasks, each governed by a sharp physical boundary, such as supported vs.\ falling, passing vs.\ blocked, or stable vs.\ collapsing.
\levelicon{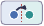}\emph{Level 1 (Detection)} tests whether a model's prediction changes across the tipping point using 300 \emph{visual minimal pairs}: pairs of nearly identical initial frames on opposite sides of the boundary whose true futures differ, which a world model must tell apart.
\levelicon{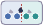}\emph{Level 2 (Selectivity)} tests whether the prediction changes \emph{only} when the true future does, adding equal-sized control perturbations on the same side of the boundary to penalize models that flip at every minor visual change.
\levelicon{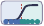}\emph{Level 3 (Localization)} tests \emph{where} and \emph{how sharply} the belief transitions by sweeping densely across the boundary to compare the predicted switch with the true physical boundary.
Levels 2 and 3 use the simulated tasks, where the boundary variable is known exactly.

We benchmark 11 video world models on FlipTest, from general video generators~\citep{google2026geminiomniflash, wan2025, wu2025hunyuanvideo, yang2025cogvideox} to physical-AI~\citep{cosmos1, cosmos3} and robotics-focused predictors~\citep{guo2026video, ma2026scaling}, evaluating over 21{,}000 generated video rollouts.
Even the strongest model, Gemini Omni Flash 1.1, averages only 18.3\% pair accuracy across real and simulated scenes, falling below the 25\% of random guessing, while on simulated tasks, \emph{every single model} performs worse than chance.
On Levels 2 and 3, models flip their predictions as often on inconsequential control steps as on decisive boundary crossings, exhibiting flat response curves across the sweeps.
Larger models, newer versions, and physics-specific training help little.

Why do video world models fail? To diagnose the bottleneck, we decompose visual prediction into perception, dynamics, and rendering~\citep{wu2017vda} and compare video models against predictors that isolate or bypass specific stages.
Bypassing pixel rendering with a vision-language model (VLM) that predicts outcomes in text already yields selective transitions at the boundary.
A coding agent that explicitly measures pixel geometry localizes the boundary with high precision, and feeding those measurements into a physics engine alters only 4 of 761 predictions.
The outcome can therefore be read from the initial frame alone, and once the state is measured, physics is the easy part.

\begin{figure}[t]
    \centering
    \includegraphics[width=\linewidth]{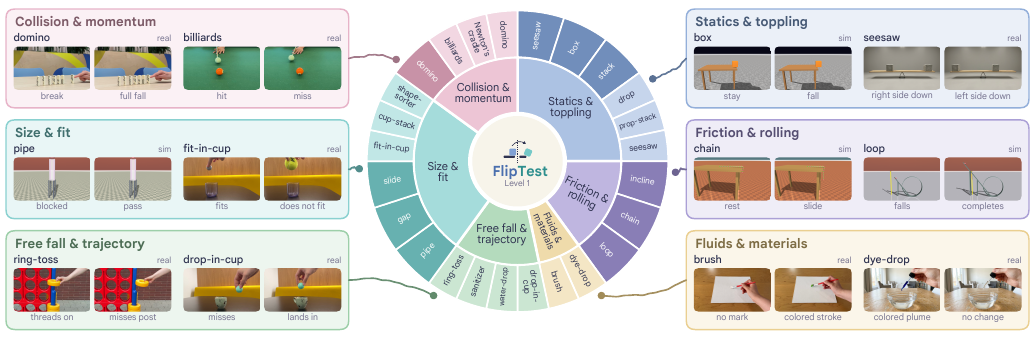}
    \caption{\textbf{FlipTest-Level 1 at a glance.}
It has 300 minimal pairs (600 initial frames) over 25 tasks (outer ring), grouped into six physical families (inner ring).
Dark wedges are simulated tasks and light wedges are real tasks, with 150 pairs and 300 initial frames each.
Each callout shows two example pairs from one family.
Each pair has two opposite outcomes.}
    \label{fig:fig2-benchmark-overview}
\end{figure}

This leaves the video model with two entangled responsibilities: reasoning about the outcome and rendering the pixels.
To isolate rendering, we provide the ground-truth outcome in the prompt.
The best video model surges from 18.3\% to 73.7\% pair accuracy, with strong gains across all models.
This implies modern video models can often \emph{draw} the right future, yet cannot deduce from the initial frame \emph{which} future will unfold, and even when told, the best model misses over a quarter of the pairs.

We show that today's video world models have learned how the future \emph{looks}, but not what \emph{decides} it.
Next-frame prediction, as currently formulated, fails to compel models to extract precise physical state from visual observations.
FlipTest makes this fundamental reasoning gap measurable, offering a rigorous, scalable yardstick to guide the next generation of physically grounded world models.

\begin{figure}[t]
    \centering
    \includegraphics[width=\linewidth]{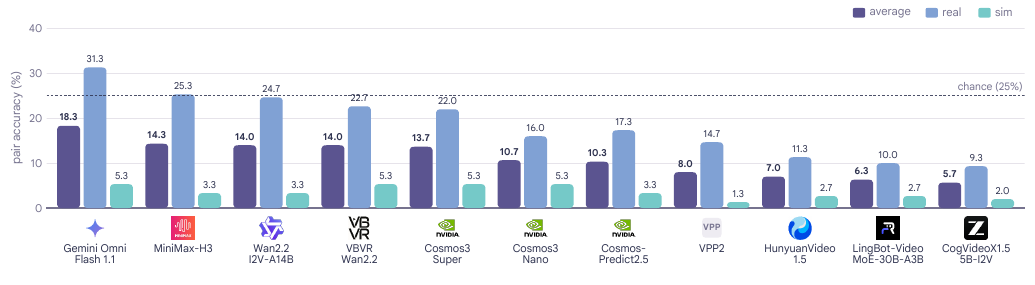}
    \caption{\textbf{Video world models rarely pass FlipTest Level 1.}
We show pair accuracy (\%) of 11 video models on real scenes, simulation, and their average.
The models are sorted by their average.
A frame counts as correct only if both judges (Gemini~3.8 Flash and Gemini~3.6 Flash) read the ground-truth outcome from its video.
A pair is solved only if both frames are correct.
The dashed line marks chance (25\%).}
    \label{fig:fig3-pair-accuracy}
\end{figure}

\section{Related Work}
\label{sec:related}

\paragraph{Physics evaluation of video and world models.}
Video world models aim to learn how the physical world evolves~\citep{cosmos1, cosmos3, sora}, and recent work improves their dynamics with predictive structure~\citep{luo2026predictive}.
Intuitive-physics benchmarks such as IntPhys~\citep{intphys}, Physion~\citep{physion}, PHYRE~\citep{phyre}, and CLEVRER~\citep{clevrer} evaluate physical prediction and reasoning in synthetic scenes, a setup \citet{garrido2025intuitive} adapt to test whether self-supervised video models~\citep{vjpea2} detect physical violations.
For generative models, VideoPhy~\citep{bansal2025videophy, videophy2} rates physical commonsense, Physics-IQ~\citep{physicsiq} compares continuations against real recordings, and WorldModelBench~\citep{worldmodelbench} scores instruction following and physical adherence.
However, existing evaluations score each scene in isolation, missing whether a model generates the exact same plausible video for two initial frames whose true futures are opposite.
FlipTest instead probes world models counterfactually across sharp physical boundaries: using visual minimal pairs, matched controls, and dense sweeps, it tests whether predicted outcomes flip precisely where the true future flips, and only there.

\paragraph{Minimal pairs and contrast sets.}
Minimal pairs isolate a single capability by holding everything else fixed.
They have exposed blind spots in grammar with BLiMP~\citep{blimp}, in vision--language composition with Winoground~\citep{winoground}, and in fine-grained visual perception with MMVP~\citep{mmvp}, and contrast sets~\citep{gardner2020contrast} use them to probe local decision boundaries.
Violation-of-expectation studies with infants~\citep{baillargeon1985object, teglas2011pure, margoni2024violation} rest on the same logic and compare closely matched events that differ in one physically relevant respect.
FlipTest brings this experimental control to world modeling.
It applies the minimal change to the \emph{initial state} and asks for the \emph{future}, so it probes the model's transition function directly.

\paragraph{Inverse graphics and physical scene understanding.}
Inferring hidden scene state from images dates back to analysis-by-synthesis~\citep{helmholtz1867, halle1962speech, yuille2006vision}.
Following this principle, Picture~\citep{kulkarni2015picture}, Neural Scene De-rendering~\citep{wu2017neural}, and Visual De-animation~\citep{wu2017vda} recover explicit object states from images and evolve them with a physics engine.
Recent work revives this modular paradigm via coding agents that represent state as code~\citep{wang2026codeasworlds, chen2026codeworldmodel, liu2026agenticwam}, whereas video world models collapse perception, dynamics, and rendering into one pixel-prediction network.
However, prior work builds predictors along either paradigm without diagnosing which stage fails at a physical boundary.
We repurpose the three-stage decomposition of Visual De-animation~\citep{wu2017vda} as a diagnostic: comparing video models against a rendering-free VLM and coding agents that measure geometry or run a physics engine isolates where the decisive state is lost.

\begin{figure}[t]
    {\includegraphics[width=\linewidth, trim=4.0 176.36 61.8 0, clip]{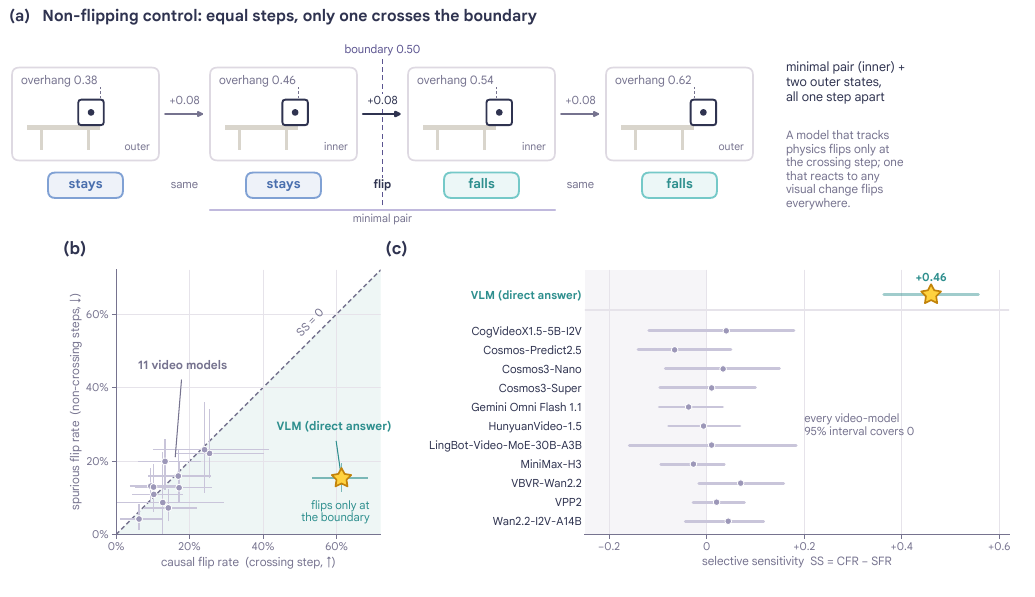}\par}
    \nointerlineskip
    \includegraphics[width=\linewidth, trim=30.2 10 11.2 110.2, clip]{fig/fig5-selectivity-controls.pdf}
    \caption{\textbf{Video models are not selectively sensitive in FlipTest Level 2.}
(a) A control group on the box task: four equally spaced states (offsets exaggerated for display), of which only the middle step crosses the boundary.
A model that reads the state flips only at the crossing step, landing below the diagonal in (b) and right of 0 in (c) like the VLM; video models sit near the diagonal, flipping about as often when the outcome should stay as when it should change.
(b) Flip rate at the crossing step vs.\ at the non-crossing steps of a control group. The shaded region marks SS $>0$.
(c) Selective sensitivity (crossing minus non-crossing flip rate) with 95\% CIs over scenes and two judges.
The VLM is Gemini 3.6 Flash answering directly from the same first frames. It flips on 61\% of crossing steps and only 15\% of non-crossing steps ($+0.46$, 95\% CI $[0.36, 0.56]$), while video models score between $-0.07$ and $+0.07$, each with a CI that covers 0.}
    \label{fig:fig5-selectivity-controls}
\end{figure}

\section{FlipTest}
\label{sec:benchmark}

How can we test whether a model's prediction is truly grounded in the present state? We first formalize this counterfactual requirement via boundary minimal pairs (\cref{sec:def}), then instantiate it through a three-level hierarchy of probes (\cref{sec:probes}), and finally define our evaluation readout and metrics (\cref{sec:readout}).

\subsection{From Prediction to Minimal Pairs}
\label{sec:def}

Let $s$ denote the underlying physical state of a scene and $y(s) \in \{0, 1\}$ denote whether a target future event occurs.
Across the physical tasks in FlipTest, the outcome is governed by a scalar or low-dimensional margin $m(s)$---such as support overhang, geometric clearance, net torque, or collision offset---whose sign determines the future:
\[
y(s) = \mathbb{I}[m(s) > 0],
\]
where $\mathbb{I}[\cdot]$ is the indicator function, equal to 1 when its argument holds and 0 otherwise.
To probe whether a model is sensitive to this threshold, we construct state pairs $(s^{-}, s^{+})$ straddling the boundary,
\[
m(s^{-}) < 0 < m(s^{+}),
\]
with both states sampled close to $m(s)=0$. Consequently, their initial visual observations are nearly identical, yet their true futures are opposite:
\[
y(s^{-}) \neq y(s^{+}).
\]
Any model that genuinely predicts the future from the present must therefore assign opposite outcomes to $s^{-}$ and $s^{+}$.
However, passing a single pair is necessary but insufficient: a stochastic or hypersensitive model could flip its prediction by chance or in reaction to any pixel change.
In \cref{sec:probes}, we therefore extend this pairwise formulation into a multi-level hierarchy that tests whether predictions flip \emph{if and only if} the true outcome changes.

\subsection{A Hierarchy of Boundary Tests}
\label{sec:probes}

To evaluate how faithfully a model tracks the physical boundary, our FlipTest samples states along $m(s)$ across 1{,}680 initial frames, organized into three difficulty levels.

\paragraph{\levelicon{1}Level 1: Pairs (Detection).}
First, does the model's prediction change across the boundary?
Each minimal pair places one state on either side of the tipping point ($m(s^-)<0<m(s^+)$).
The two initial frames share the exact same scene, objects, and camera, differing only in a single outcome-controlling variable that flips the ground-truth future (\eg stay vs.\ fall, pass vs.\ blocked).
Because the paired frames are visually matched, solving a pair requires extracting the decisive physical quantity, such as overhang, clearance, lever arm, or offset, directly from image geometry.
Level~1 forms our core benchmark (\cref{sec:pairs}), comprising \textbf{300} minimal pairs (\textbf{600} initial frames) across \textbf{25} tasks in six physical families (\cref{fig:fig2-benchmark-overview}): statics and toppling, size and fit, friction and rolling, collision and momentum, free fall and trajectory, and fluids and materials.
These are split evenly into 10~simulated tasks (15 pairs each; 300 images) and 15~real-world tasks (10 pairs each; 300 images).

Specifically, the 10 simulated tasks are built in MuJoCo~\citep{todorov2012mujoco}.
Within each pair, the two states sit at symmetric margins $\pm\delta$ around the analytic or empirically calibrated threshold.
Across pairs, $\delta$ varies from knife-edge to obvious margins, while camera viewpoints and visual themes vary across pairs but remain fixed within each pair.
Crucially, camera placement depends only on static scene geometry, ensuring that framing never leaks the outcome.

Complementing simulation, the 15 real-world tasks capture everyday scenes near empirical physical boundaries.
Of these 300 frames, 59 come from \emph{captured footage} with recorded physical outcomes, while 241 are \emph{counterfactual edits} of real photographs or \emph{staged synthetic stills}.
Each paired frame is constructed via a minimal edit that alters only the boundary variable (\eg sliding an object slightly further past a table edge), fixing the ground-truth label by construction while keeping the pair pixel-matched elsewhere.
\Cref{app:tasks} provides full construction details for all 25 tasks (\cref{tab:sim-tasks,tab:real-tasks}).

\paragraph{\levelicon{2}Level 2: Controls (Selectivity).}
Passing Level~1 shows sensitivity, but a model could still flip indiscriminately at every visible change.
To test selectivity, we augment each of the 10 simulated tasks with 15 control groups (600 frames total), where each group consists of four equally spaced states at margins $-3\delta, -\delta, +\delta, +3\delta$ (\cref{fig:fig5-selectivity-controls}a).
Only the middle step ($-\delta \to +\delta$) crosses the physical boundary; the two outer steps apply the exact same spatial perturbation and visual shift while leaving the true outcome unchanged.
A model that genuinely tracks the physical state must flip on the middle step while holding steady on the outer steps (\cref{sec:controls}).

\begin{figure}[t]
    \centering
    \includegraphics[width=\linewidth, trim=0 17 0 0, clip]{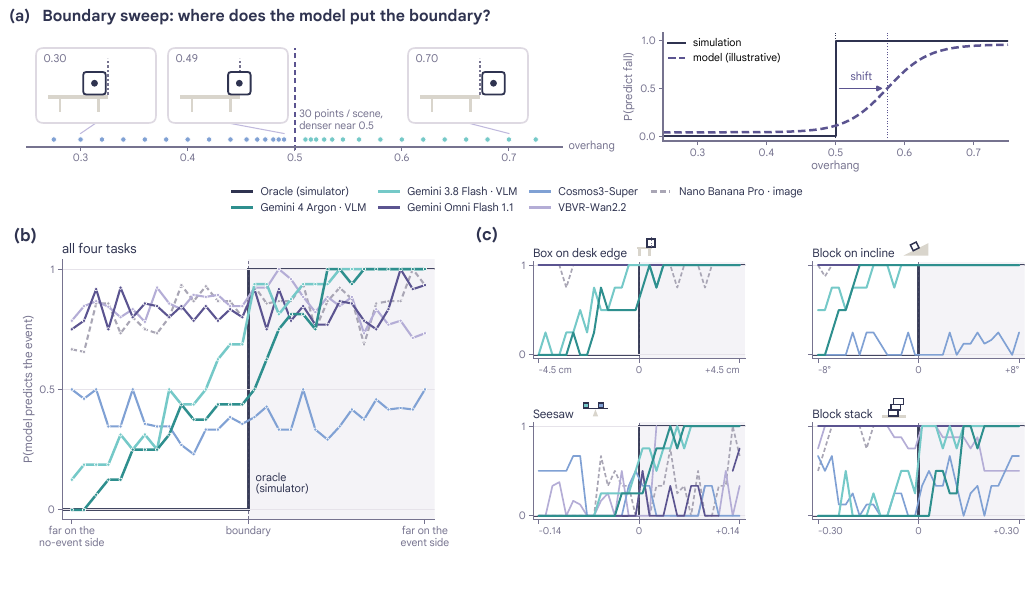}
    \caption{\textbf{Video models do not see the boundary in FlipTest Level 3.}
(a) A sweep of 30 states per scene across the boundary, densest near it; a model's $P(\mathrm{event})$ curve shows where it places the boundary and how sharply it switches (model curve illustrative).
$P(\mathrm{event})$ along the sweep is shown (b) pooled and (c) per task (4 scenes $\times$ 30 states).
A model that reads the state should follow the oracle, which jumps from 0 to 1 at the boundary.
While VLMs bend toward the oracle step at $m=0$, video models remain flat.}
    \label{fig:fig6-boundary-sweep}
\end{figure}

\paragraph{\levelicon{3}Level 3: Sweeps (Localization).}
Finally, even when a model flips selectively, \emph{where} and \emph{how sharply} does its belief transition?
Across four representative simulated tasks (box at a desk edge, block on an incline, seesaw, and block stack) with four scenes each, we densely sample 30 states per scene spanning both sides of the threshold, concentrated near $m=0$ (\cref{fig:fig6-boundary-sweep}a).
Across these 480 frames, the ground-truth outcome transitions as a sharp step function at $m=0$, allowing us to measure whether the model's transition is centered and sharp (\cref{sec:boundary}).

\subsection{Outcome Readout and Metrics}
\label{sec:readout}
\label{sec:metrics}

To evaluate diverse architectures on equal footing, every model receives only the clean initial frame and a neutral prompt stating the setup and possible outcomes, without hinting at the underlying decision rule.
While VLMs predict the outcome label directly in text from the initial frame, video world models generate a video rollout whose outcome must be read out.
To do so reliably, we employ two independent VLM judges, Gemini 3.6 Flash~\citep{google2026gemini36flash} and Gemini 3.8 Flash~\citep{google2026gemini38flash}, to classify the event depicted in each generated video (or flag it as \texttt{unclear}).
Crucially, the judges evaluate only \emph{what the generated video shows}, regardless of whether that motion is physically realistic.
On Level~1 pairs, a rollout is scored as correct only if both judges agree on the ground-truth outcome; inter-judge disagreements and \texttt{unclear} rollouts are scored as incorrect, since a world model useful for planning must commit to a clear future.
For Level~2 controls and Level~3 sweeps, we retain both judges' labels to estimate transition statistics (\cref{app:judge} details judge prompts, settings, and inter-annotator agreement).

Corresponding to the three levels of FlipTest, we evaluate models with three complementary metrics:

\paragraph{Pair accuracy (Level 1).}
A minimal pair is scored as correct if and only if the model predicts the correct outcome for both frames.
Under this strict paired metric, random guessing on a binary outcome yields 25\%, whereas a constant predictor that ignores the visual perturbation achieves 0\%.

\paragraph{Selective sensitivity (Level 2).}
On the four-state control groups, we compare the \emph{crossing flip rate} (CFR, on the $-\delta \to +\delta$ step) against the \emph{same-side flip rate} (SFR, averaged over the $-3\delta \to -\delta$ and $+\delta \to +3\delta$ steps).
We define selective sensitivity as $\mathrm{SS}=\mathrm{CFR}-\mathrm{SFR}$, which equals zero for both constant predictors and random-flip models, and is positive only when predictions change preferentially across the true boundary.
We report 95\% bootstrap confidence intervals over 2{,}000 draws by resampling control groups and sampling one judge per frame.

\paragraph{Response curve (Level 3).}
On dense sweeps, we plot the psychometric curve $P(\mathrm{event}\mid m)$, the fraction of rollouts depicting the target event at each margin $m$.
Whereas the ground-truth simulator exhibits a crisp step function at $m=0$, a flat, shifted, or gradual curve immediately diagnoses a missing, biased, or blurred physical boundary.

\section{Video Models Fail to Learn Physical State}
\label{sec:main_results}

\paragraph{Models.}
We evaluate 11 video world models across three categories: general video generators, including Gemini Omni Flash 1.1~\citep{google2026geminiomniflash}, MiniMax-H3~\citep{minimax2026h3}, Wan2.2-I2V-A14B~\citep{wan2025} alongside its reasoning-tuned variant VBVR-Wan2.2~\citep{wang2026very}, HunyuanVideo-1.5~\citep{wu2025hunyuanvideo}, and CogVideoX1.5-5B-I2V~\citep{yang2025cogvideox}; physical-AI foundation models, comprising Cosmos-Predict2.5~\citep{cosmos1}, Cosmos3-Nano~\citep{cosmos3}, and Cosmos3-Super~\citep{cosmos3}; and robotics-focused predictors, namely VPP2~\citep{guo2026video} and LingBot-Video-MoE-30B-A3B~\citep{ma2026scaling}.
Each model generates a single video rollout per initial frame.

\subsection{\levelicon{1}Video Models Rarely Solve Minimal Pairs}
\label{sec:pairs}

If a video model truly tracks the underlying physical state, the two frames of a minimal pair must lead to opposite futures.
\Cref{fig:fig3-pair-accuracy} reveals that they rarely do.
On simulated tasks, no model exceeds 5.3\% pair accuracy (a ceiling shared by four models), falling far below the 25\% expected from random guessing.
On real-world scenes, the strongest model, Gemini Omni Flash 1.1, reaches only 31.3\%, followed by MiniMax-H3 (25.3\%) and Wan2.2-I2V-A14B (24.7\%).
Only the top two models surpass chance, while CogVideoX1.5-5B-I2V drops to 9.3\%.
Averaged across both splits, overall accuracy spans a meager 5.7\% to 18.3\%.

\begin{figure}[t]
    \centering
    \includegraphics[width=\linewidth]{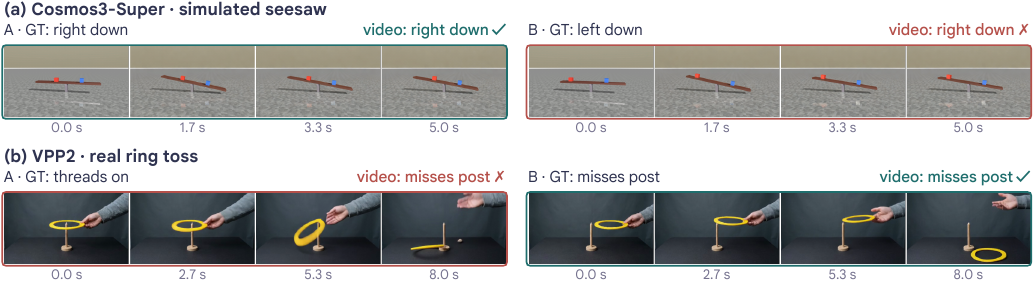}
    \caption{\textbf{Failure mode: one future for both frames.}
Four evenly spaced frames of the generated videos for two minimal pairs, (a) Cosmos3-Super on the simulated seesaw and (b) VPP2 on real ring toss, labeled with the outcome both judges read.
Each model shows the same outcome for both frames of its pair, so one of the two videos is wrong (red border).}
    \label{fig:same-future}
\end{figure}

Why do video models perform worse than random guessing?
The primary culprit is mode collapse across the boundary: models overwhelmingly generate the \emph{exact same} outcome for both frames of a pair, guaranteeing that at least one rollout is wrong (\cref{fig:same-future}).
Pooled across all models on clearly judged pairs, predictions remain identical across 72\% of real pairs and 89\% of simulated pairs.
Compounding this insensitivity, generated rollouts frequently violate basic physics.
In roughly 14\% of simulated evaluations, both judges flagged at least one video as \texttt{unclear}, most notably for CogVideoX1.5-5B-I2V and LingBot-Video-MoE-30B-A3B, as objects spontaneously duplicated, deformed, or launched into midair (\cref{app:results}).

\paragraph{Real scenes offer semantic shortcuts.}
Every model scores 7 to 26 percentage points higher on real scenes than on simulation.
Real photographs often contain familiar visual priors that correlate with the outcome, such as a cup visibly overhanging a table edge or a ring clearly wider than its post.
By contrast, our simulated split holds visual appearance strictly constant within each pair and perturbs only the continuous boundary variable, making it a purer and stricter test of physical state reading.

\paragraph{Scale, newer versions, and specialized training help little.}
Scaling up model size, upgrading architectures, or applying domain-specific post-training yields negligible gains.
The flagship physical-AI model Cosmos3-Super (13.7\%) performs on par with the open general-purpose generator Wan2.2-I2V-A14B (14.0\%).
Within the Cosmos family, scaling from Cosmos-Predict2.5 (10.3\%) and Cosmos3-Nano (10.7\%) to Cosmos3-Super (13.7\%) brings only marginal gains, and all three remain capped at or below 5.3\% on simulation.
Similarly, reasoning-tuned VBVR-Wan2.2 (14.0\%) matches its base model Wan2.2-I2V-A14B (14.0\%) exactly, while robotics-focused predictors VPP2 (8.0\%) and LingBot-Video-MoE-30B-A3B (6.3\%) lag in the bottom half of the leaderboard.
\begin{keyidea}
\finding{1}Video world models rarely predict opposite futures for minimal pairs, performing worse than chance across all models on simulation.\par
\finding{2}Larger scale, newer model generations, physics-specific pretraining, and reasoning tuning yield \mbox{negligible~gains.}
\end{keyidea}

\subsection{\levelicon{2}Video Models Are Not Selectively Sensitive}
\label{sec:controls}

Failing Level~1 minimal pairs could stem from two distinct behaviors: either a model is completely rigid and never flips its prediction, or it flips erratically at any pixel change regardless of the physical boundary.
Our Level~2 controls disentangle these two failure modes.
Plotting the crossing flip rate against the same-side flip rate (\cref{fig:fig5-selectivity-controls}b) shows that all 11 video models cluster tightly along the diagonal: when they do change their prediction, they flip just as often on inconsequential control steps as on decisive boundary crossings.
Consequently, selective sensitivity (\cref{fig:fig5-selectivity-controls}c) hovers near zero for every model, ranging from $-0.07$ (Cosmos-Predict2.5) to $+0.07$ (VBVR-Wan2.2), with every 95\% confidence interval straddling zero.
\begin{keyidea}
\finding{3}Models lack selective sensitivity, flipping their predictions just as often on harmless control steps as across true physical boundaries.
\end{keyidea}

\subsection{\levelicon{3}Video Models Do Not Locate the Boundary}
\label{sec:boundary}

While Level~2 probes a single crossing step, Level~3 sweeps test whether a model merely misplaces or blurs the boundary by tracking its full response across the threshold.
We compare three video models (Gemini Omni Flash 1.1, Cosmos3-Super, VBVR-Wan2.2), the single-frame generator Nano Banana Pro~\citep{google2025gemini3proimage}, and two direct VLMs (Gemini 4 Argon, Gemini 3.8 Flash).
Plotting $P(\mathrm{event}\mid m)$ against the oracle step (\cref{fig:fig6-boundary-sweep}b,c) shows that all pixel generators produce flat curves, with cross-boundary probability gaps of only $0.00$ to $+0.06$.
Video predictions ignore the margin even at the sweep extremes, defaulting instead to a constant, task-specific bias in each panel (\cref{fig:fig6-boundary-sweep}c).

By contrast, both VLMs follow clear S-shaped transitions with gaps of $+0.62$ (Gemini 4 Argon) and $+0.58$ (Gemini 3.8 Flash), though their curves rise before $m=0$, over-predicting the event near the threshold.
Thus, whereas video models fail to register the boundary at all, VLMs detect it with a systematic shift.
If a VLM can read the boundary from the initial frame, where along the generation pipeline does a video model lose it (\cref{sec:failure})?
\begin{keyidea}
\finding{4}Video world models remain flat across the entire boundary sweep, whereas VLMs transition clearly but prematurely.
\end{keyidea}

\section{Why Do Video World Models Fail?}
\label{sec:failure}

\begin{figure}[t]
    \centering
    \includegraphics[width=\linewidth]{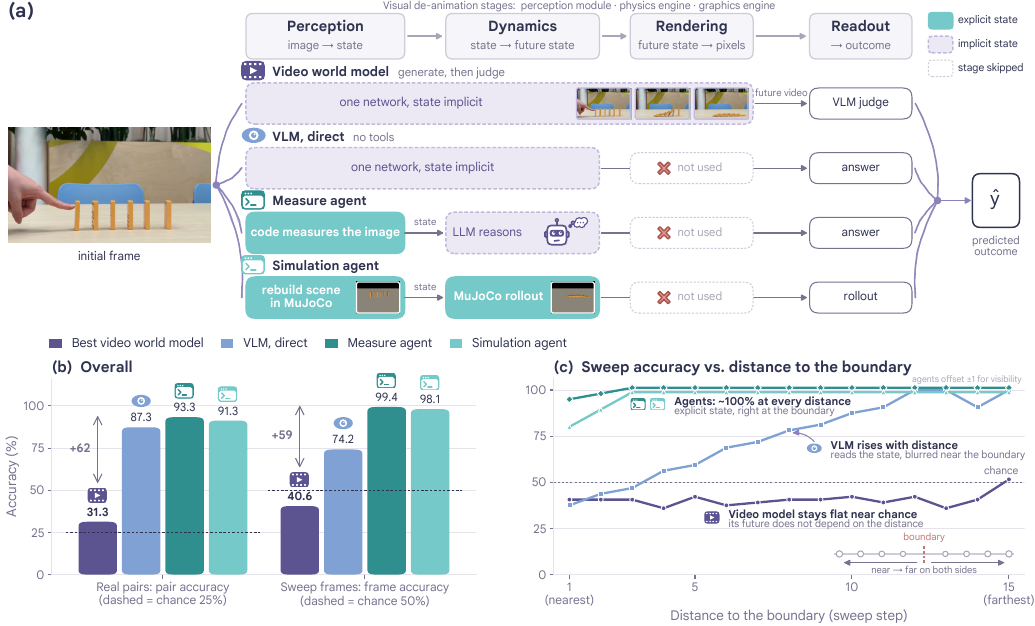}
    \caption{\textbf{Making the state explicit closes the gap.}
(a) Predictors mapped onto the three stages of Visual De-animation~\citep{wu2017vda}, with every pipeline terminating in a binary outcome label scored against ground truth.
(b) Pair accuracy on the 150 real pairs and frame accuracy on the 480 sweep frames, comparing the best video world model, the direct VLM, the Measure agent, and the Simulation agent against chance (dashed lines: 25\% for pairs, 50\% for frames).
(c) Sweep accuracy as a function of distance from the boundary, pooled over four tasks, four scenes, and both boundary sides (32 frames per step; invalid agent runs count as wrong).}
    \label{fig:fig7-agent-pipeline-gap}
\end{figure}

To uncover why video world models fail to reason across physical boundaries, we must dissect how visual prediction works under the hood.
Predicting a physical outcome from an image requires three stages: perceiving the state, evolving the dynamics, and, for pixel generators, rendering the future.
Following Visual De-animation (VDA)~\citep{wu2017vda}, which separates these stages into a perception module, a physics engine, and a graphics engine, we map predictors onto this pipeline to isolate where boundary reasoning breaks down (\cref{fig:fig7-agent-pipeline-gap}a).
Whereas a video world model entangles all three stages inside a single network, alternative predictors isolate or bypass specific stages: a direct VLM bypasses rendering to predict outcomes in text; a coding agent makes perception explicit by measuring image geometry with code; and a simulation agent makes dynamics explicit by evolving those measurements in a physics engine.
Conversely, providing the ground-truth outcome in the video model's prompt isolates rendering on its own.
Evaluating every predictor on the exact same binary outcome label $\hat{y}$ pinpoints which stage loses the decisive physical state.

\paragraph{The frame contains the decisive information.}
Bypassing pixel rendering with a VLM immediately confirms that the boundary is visually recoverable from the initial frame.
On Level~2 controls, Gemini 3.6 Flash flips its prediction on 61\% of crossing steps versus only 15\% of same-side steps, achieving a selective sensitivity of $+0.46$ (95\% CI $[0.36, 0.56]$; \cref{fig:fig5-selectivity-controls}), whereas every video model hovers around zero.
Similarly, on Level~3 sweeps, Gemini 4 Argon and Gemini 3.8 Flash trace clear S-shaped transitions where all video models remain flat (\cref{fig:fig6-boundary-sweep}).
Initial frames that video models treat as interchangeable are readily distinguished by a model that reads them directly, though the VLMs' early, gradual transitions indicate imprecise perceptual estimation near the threshold.

\begin{figure}[t]
    \centering
    \includegraphics[width=\linewidth]{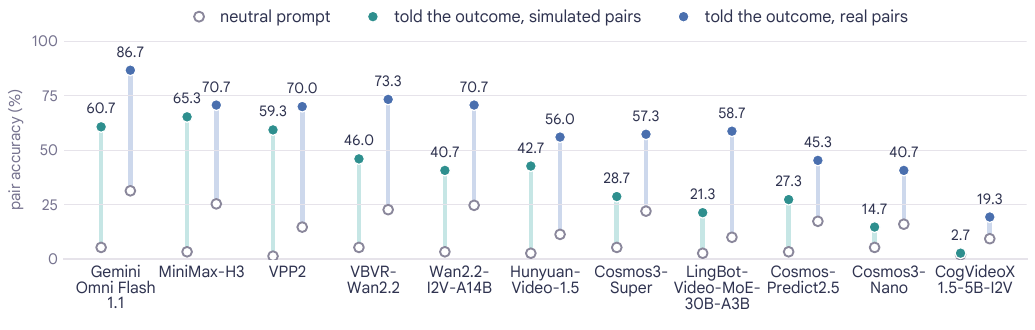}
    \caption{\textbf{Told the outcome, every video model improves.}
Pair accuracy (\%) with the neutral prompt (hollow, as in \cref{fig:fig3-pair-accuracy}) and with the ground-truth outcome stated in the prompt (filled; teal for simulated pairs, blue for real pairs).
A pair counts only when both videos show the told outcome, scored by Gemini 3.8 Flash with \texttt{unclear} rollouts counted as wrong.}
    \label{fig:told-outcome}
\end{figure}

\paragraph{Told the outcome, video models improve but still fail.}
If the state is visible in the frame, do video models fail because they cannot deduce the outcome, or because they cannot render it?
To isolate rendering, we modify each prompt to state the ground-truth outcome explicitly and regenerate all pair rollouts across all models (\cref{fig:told-outcome}; \cref{app:results}).
Relieved of deciding the outcome, Gemini Omni Flash 1.1 surges from 18.3\% to 73.7\% pair accuracy, with substantial gains across every model.
This gap confirms that video models largely know how to \emph{draw} the future, but fail to infer from the initial frame \emph{which} future will unfold.
Even so, conditional rendering remains imperfect: the strongest model still misses over a quarter of the pairs, performing better on real scenes (86.7\%) than on simulated ones (60.7\%).

\paragraph{Measuring the state recovers the boundary.}
To see if explicit perceptual measurement can eliminate the remaining boundary errors seen in direct VLMs, we evaluate three predictors built on Gemini 4 Argon with the exact same neutral prompt and clean initial frame, differing only in how they process the state (\cref{fig:fig7-agent-pipeline-gap}a):
(1)~\textbf{VLM (direct)}, answering from the image in a single forward pass without tools;
(2)~\textbf{Measure agent}, writing and executing Python image-processing scripts (NumPy, SciPy, scikit-image) to measure pixel geometry before answering; and
(3)~\textbf{Simulation agent}, further reconstructing the measured scene in MuJoCo~\citep{todorov2012mujoco} and reading the outcome from an audited offline rollout (\cref{app:agent}).
Evaluated on real pairs and sweep frames (\cref{fig:fig7-agent-pipeline-gap}b,c), the direct VLM already outperforms the best video model on real pairs (87.3\% vs.\ 31.3\%; \cref{fig:fig7-agent-pipeline-gap}b) and reaches 74.2\% frame accuracy on sweeps, but its errors concentrate near the tipping point: sweep accuracy drops to 38\% at steps adjacent to $m=0$ and only exceeds 90\% eleven steps away (\cref{fig:fig7-agent-pipeline-gap}c).
Making perception explicit via the Measure agent eliminates nearly all of this boundary blur, boosting accuracy to 93.3\% on real pairs and over 99\% across the sweeps, with only three errors encountered across all sweep frames (\cref{app:agent-analysis}).

\paragraph{Simulation adds little to measurement.}
Adding an explicit physics engine on top of code-based measurement yields virtually no further gain.
Across evaluation frames, the Simulation agent successfully builds and executes MuJoCo scenes in the vast majority of cases, solving over 91\% of real pairs and 98\% of sweep frames to match the Measure agent.
Indeed, the Simulation agent's pre-simulation estimate from code measurements alone already reaches 96.0\% on real frames and 97.5\% on sweep frames; running the physical rollout alters the final prediction in less than 1\% of runs (\cref{app:agent-analysis}), while requiring roughly ten minutes per frame (\cref{app:agent}).
\begin{keyidea}
\finding{5}State measurement from the initial frame solves over 93\% of real pairs, making explicit physics simulation redundant.\par
\finding{6}Video models fail primarily at reading state and secondarily at rendering, missing over a quarter of pairs even when told the ground-truth outcome.
\end{keyidea}

\section{Conclusion}
\label{sec:conclusion}

We introduced FlipTest, a benchmark of minimal pairs that asks whether a world model changes its prediction when a small change to the present flips what happens next.
Current video world models mostly do not.
The best solves 31.3\% of real pairs and 5.3\% of simulated pairs, and models generate the same outcome for both frames in most pairs.
Video models also flip as often within one side of the boundary as across it, and their predictions stay flat across the boundary sweep.
Scaling, newer versions, physics-specific pretraining, and reasoning tuning barely help.

Our failure analysis traces this gap mainly to reading the physical state and secondarily to rendering.
From the same first frame, VLMs track the boundary and a coding agent that measures the state solves over 93\% of real pairs, while video world models told the ground-truth outcome still miss over a quarter of them.

This gap matters in the agentic era.
Models like Gemini 4 Carbon~\citep{google2026gemini4carbon} and Fable 5.2~\citep{anthropic2026fable52} point toward generalist physical agents that operate in real environments, where trial and error is slow, costly, and fragile.
World models promise a virtual sandbox where such agents can simulate thousands of trials and receive physically grounded feedback before they act.
In this role, basic causality is essential, and current video world models do not yet provide it.

Our results point toward a modular, agentic blueprint for future world simulators.
An agent estimates the physical state from the scene, calls a dedicated simulator or numerical solver when needed, and passes the predicted trajectory to a video model that renders the motion.
This division builds on inverse graphics~\citep{wu2017vda, battaglia2013simulation} and analysis-by-synthesis~\citep{yuille2006vision}, as well as recent systems that use code as world representations~\citep{chen2026codeworldmodel, liu2026agenticwam, wang2026codeasworlds}.
Agents already find the correct outcome, so the open challenge is to have video models render a given outcome faithfully without overriding the physics.
We discuss limitations in \cref{app:limitations}.
We hope FlipTest helps the community rethink world modeling as pure video prediction and build world models that learn both how the future looks and what decides it.

\bibliographystyle{abbrvnat}
\bibliography{references}

\begin{thebibliography}{59}
\providecommand{\natexlab}[1]{#1}
\providecommand{\url}[1]{\texttt{#1}}
\expandafter\ifx\csname urlstyle\endcsname\relax
  \providecommand{\doi}[1]{doi: #1}\else
  \providecommand{\doi}{doi: \begingroup \urlstyle{rm}\Url}\fi

\bibitem[Agarwal et~al.(2025)Agarwal, Ali, Bala, Balaji, Barker, Cai,
  Chattopadhyay, Chen, Cui, Ding, et~al.]{cosmos1}
N.~Agarwal, A.~Ali, M.~Bala, Y.~Balaji, E.~Barker, T.~Cai, P.~Chattopadhyay,
  Y.~Chen, Y.~Cui, Y.~Ding, et~al.
\newblock Cosmos world foundation model platform for {Physical AI}.
\newblock \emph{arXiv preprint arXiv:2501.03575}, 2025.

\bibitem[Agarwal et~al.(2026)Agarwal, Ali, Allen, Antolini, Aubame, Azzolini,
  Bai, Bala, Balaji, Bapst, et~al.]{cosmos3}
N.~Agarwal, A.~Ali, J.~Allen, M.~Antolini, A.~Aubame, A.~Azzolini, J.~Bai,
  M.~Bala, Y.~Balaji, J.~Bapst, et~al.
\newblock Cosmos 3: Omnimodal world models for {Physical AI}.
\newblock \emph{arXiv preprint arXiv:2606.02800}, 2026.

\bibitem[{Anthropic}(2026)]{anthropic2026fable52}
{Anthropic}.
\newblock {Claude Fable 5.2} and {Mythos 5.2} system card, Nov. 2026.
\newblock URL
  \url{https://www.anthropic.com/document/claude-fable-5-2-mythos-5-2-system-card}.

\bibitem[Assran et~al.(2025)Assran, Bardes, Fan, Garrido, Howes, Muckley,
  Rizvi, Roberts, Sinha, Zholus, et~al.]{vjpea2}
M.~Assran, A.~Bardes, D.~Fan, Q.~Garrido, R.~Howes, M.~Muckley, A.~Rizvi,
  C.~Roberts, K.~Sinha, A.~Zholus, et~al.
\newblock {V-JEPA} 2: Self-supervised video models enable understanding,
  prediction and planning.
\newblock \emph{arXiv preprint arXiv:2506.09985}, 2025.

\bibitem[Baillargeon(2004)]{baillargeon2004infants}
R.~Baillargeon.
\newblock Infants' physical world.
\newblock \emph{Current directions in psychological science}, 13\penalty0
  (3):\penalty0 89--94, 2004.

\bibitem[Baillargeon et~al.(1985)Baillargeon, Spelke, and
  Wasserman]{baillargeon1985object}
R.~Baillargeon, E.~S. Spelke, and S.~Wasserman.
\newblock Object permanence in five-month-old infants.
\newblock \emph{Cognition}, 20\penalty0 (3):\penalty0 191--208, 1985.

\bibitem[Baillargeon et~al.(1992)Baillargeon, Needham, and
  DeVos]{baillargeon1992support}
R.~Baillargeon, A.~Needham, and J.~DeVos.
\newblock The development of young infants' intuitions about support.
\newblock \emph{Early development and parenting}, 1\penalty0 (2):\penalty0
  69--78, 1992.

\bibitem[Bakhtin et~al.(2019)Bakhtin, van~der Maaten, Johnson, Gustafson, and
  Girshick]{phyre}
A.~Bakhtin, L.~van~der Maaten, J.~Johnson, L.~Gustafson, and R.~Girshick.
\newblock Phyre: A new benchmark for physical reasoning.
\newblock \emph{Advances in Neural Information Processing Systems}, 32, 2019.

\bibitem[Bansal et~al.(2025)Bansal, Lin, Xie, Zong, Yarom, Bitton, Jiang, Sun,
  Chang, and Grover]{bansal2025videophy}
H.~Bansal, Z.~Lin, T.~Xie, Z.~Zong, M.~Yarom, Y.~Bitton, C.~Jiang, Y.~Sun,
  K.-W. Chang, and A.~Grover.
\newblock {VideoPhy}: Evaluating physical commonsense for video generation.
\newblock In \emph{International Conference on Learning Representations},
  volume 2025, pages 102075--102121, 2025.

\bibitem[Bansal et~al.(2026)Bansal, Peng, Bitton, Goldenberg, Grover, and
  Chang]{videophy2}
H.~Bansal, C.~Peng, Y.~Bitton, R.~Goldenberg, A.~Grover, and K.-W. Chang.
\newblock {VideoPhy-2}: A challenging action-centric physical commonsense
  evaluation in video generation.
\newblock In \emph{International Conference on Learning Representations},
  volume 2026, pages 118456--118470, 2026.

\bibitem[Battaglia et~al.(2013)Battaglia, Hamrick, and
  Tenenbaum]{battaglia2013simulation}
P.~W. Battaglia, J.~B. Hamrick, and J.~B. Tenenbaum.
\newblock Simulation as an engine of physical scene understanding.
\newblock \emph{Proceedings of the national academy of sciences}, 110\penalty0
  (45):\penalty0 18327--18332, 2013.

\bibitem[Bear et~al.(2021)Bear, Wang, Mrowca, Binder, Tung, Pramod, Holdaway,
  Tao, Smith, Sun, et~al.]{physion}
D.~M. Bear, E.~Wang, D.~Mrowca, F.~J. Binder, H.-Y.~F. Tung, R.~Pramod,
  C.~Holdaway, S.~Tao, K.~Smith, F.-Y. Sun, et~al.
\newblock Physion: Evaluating physical prediction from vision in humans and
  machines.
\newblock \emph{arXiv preprint arXiv:2106.08261}, 2021.

\bibitem[Chen et~al.(2026)Chen, Lin, and Zhang]{chen2026codeworldmodel}
Y.~Chen, G.~Lin, and C.~Zhang.
\newblock Code world model: Coding agent as world brain.
\newblock \emph{arXiv preprint arXiv:2608.25927}, 2026.

\bibitem[Craik(1967)]{craik1967nature}
K.~J.~W. Craik.
\newblock \emph{The nature of explanation}, volume 445.
\newblock CUP Archive, 1967.

\bibitem[Fischer et~al.(2016)Fischer, Mikhael, Tenenbaum, and
  Kanwisher]{fischer2016functional}
J.~Fischer, J.~G. Mikhael, J.~B. Tenenbaum, and N.~Kanwisher.
\newblock Functional neuroanatomy of intuitive physical inference.
\newblock \emph{Proceedings of the national academy of sciences}, 113\penalty0
  (34):\penalty0 E5072--E5081, 2016.

\bibitem[Gardner et~al.(2020)Gardner, Artzi, Basmov, Berant, Bogin, Chen,
  Dasigi, Dua, Elazar, Gottumukkala, et~al.]{gardner2020contrast}
M.~Gardner, Y.~Artzi, V.~Basmov, J.~Berant, B.~Bogin, S.~Chen, P.~Dasigi,
  D.~Dua, Y.~Elazar, A.~Gottumukkala, et~al.
\newblock Evaluating models’ local decision boundaries via contrast sets.
\newblock In \emph{Findings of the Association for Computational Linguistics:
  EMNLP 2020}, pages 1307--1323, 2020.

\bibitem[Garrido et~al.(2025)Garrido, Ballas, Assran, Bardes, Najman, Rabbat,
  Dupoux, and LeCun]{garrido2025intuitive}
Q.~Garrido, N.~Ballas, M.~Assran, A.~Bardes, L.~Najman, M.~Rabbat, E.~Dupoux,
  and Y.~LeCun.
\newblock Intuitive physics understanding emerges from self-supervised
  pretraining on natural videos.
\newblock \emph{arXiv preprint arXiv:2502.11831}, 2025.

\bibitem[{Google DeepMind}(2025)]{google2025gemini3proimage}
{Google DeepMind}.
\newblock {Gemini 3 Pro Image}: Model card, Nov. 2025.
\newblock URL
  \url{https://deepmind.google/models/model-cards/gemini-3-pro-image/}.

\bibitem[{Google DeepMind}(2026{\natexlab{a}})]{google2026gemini36flash}
{Google DeepMind}.
\newblock {Gemini 3.6 Flash}: Model card, July 2026{\natexlab{a}}.
\newblock URL
  \url{https://deepmind.google/models/model-cards/gemini-3-6-flash/}.

\bibitem[{Google DeepMind}(2026{\natexlab{b}})]{google2026gemini38flash}
{Google DeepMind}.
\newblock {Gemini 3.8 Flash}: Model card, Sept. 2026{\natexlab{b}}.
\newblock URL
  \url{https://deepmind.google/models/model-cards/gemini-3-8-flash/}.

\bibitem[{Google DeepMind}(2026{\natexlab{c}})]{google2026gemini4carbon}
{Google DeepMind}.
\newblock {Gemini 4 Carbon}: Model card, Nov. 2026{\natexlab{c}}.
\newblock URL
  \url{https://deepmind.google/models/model-cards/gemini-4-carbon/}.

\bibitem[{Google DeepMind}(2026{\natexlab{d}})]{google2026geminiomniflash}
{Google DeepMind}.
\newblock {Gemini Omni Flash}: Model card, Aug. 2026{\natexlab{d}}.
\newblock URL
  \url{https://deepmind.google/models/model-cards/gemini-omni-flash/}.

\bibitem[Guo et~al.(2026)Guo, Yan, Zhong, Zhang, Yang, Lu, Chen, Wang, Deng,
  Yang, et~al.]{guo2026video}
Y.~Guo, H.~Yan, Z.~Zhong, Z.~Zhang, Q.~Yang, Q.~Lu, X.~Chen, Y.-J. Wang,
  S.~Deng, C.~Yang, et~al.
\newblock Video prediction policy 2: Predict better, act better.
\newblock \emph{arXiv preprint arXiv:2610.10270}, 2026.

\bibitem[Ha and Schmidhuber(2018)]{ha2018world}
D.~Ha and J.~Schmidhuber.
\newblock World models.
\newblock \emph{arXiv preprint arXiv:1803.10122}, 2\penalty0 (3):\penalty0 440,
  2018.

\bibitem[Hafner et~al.(2025)Hafner, Pasukonis, Ba, and
  Lillicrap]{hafner2025mastering}
D.~Hafner, J.~Pasukonis, J.~Ba, and T.~Lillicrap.
\newblock Mastering diverse control tasks through world models.
\newblock \emph{Nature}, 640\penalty0 (8059):\penalty0 647--653, 2025.

\bibitem[Halle and Stevens(1962)]{halle1962speech}
M.~Halle and K.~Stevens.
\newblock Speech recognition: A model and a program for research.
\newblock \emph{IRE transactions on information theory}, 8\penalty0
  (2):\penalty0 155--159, 1962.

\bibitem[Hegarty(2004)]{hegarty2004mechanical}
M.~Hegarty.
\newblock Mechanical reasoning by mental simulation.
\newblock \emph{Trends in cognitive sciences}, 8\penalty0 (6):\penalty0
  280--285, 2004.

\bibitem[Jiang et~al.(2026)Jiang, Masotto, and Sun]{waymo2026worldmodel}
C.~M. Jiang, X.~Masotto, and B.~Sun.
\newblock The {Waymo} world model: A new frontier for autonomous driving
  simulation.
\newblock Waymo blog, Feb. 2026.
\newblock URL
  \url{https://waymo.com/blog/2026/02/the-waymo-world-model-a-new-frontier-for-autonomous-driving-simulation/}.

\bibitem[Kubricht et~al.(2017)Kubricht, Holyoak, and Lu]{kubricht2017intuitive}
J.~R. Kubricht, K.~J. Holyoak, and H.~Lu.
\newblock Intuitive physics: Current research and controversies.
\newblock \emph{Trends in cognitive sciences}, 21\penalty0 (10):\penalty0
  749--759, 2017.

\bibitem[Kulkarni et~al.(2015)Kulkarni, Kohli, Tenenbaum, and
  Mansinghka]{kulkarni2015picture}
T.~D. Kulkarni, P.~Kohli, J.~B. Tenenbaum, and V.~Mansinghka.
\newblock Picture: A probabilistic programming language for scene perception.
\newblock In \emph{Proceedings of the ieee conference on computer vision and
  pattern recognition}, pages 4390--4399, 2015.

\bibitem[Li et~al.(2026)Li, Fang, Chen, Yang, Cao, Wong, Luo, Wang, Yin,
  Gonzalez, et~al.]{worldmodelbench}
D.~Li, Y.~Fang, Y.~Chen, S.~Yang, S.~Cao, J.~Wong, M.~Luo, X.~Wang, H.~Yin,
  J.~Gonzalez, et~al.
\newblock Worldmodelbench: Judging video generation models as world models.
\newblock \emph{Advances in Neural Information Processing Systems}, 38, 2026.

\bibitem[Liu et~al.(2026)Liu, Niu, Zhang, Shao, Wang, Zheng, Sharma, Lu, Liu,
  Zhang, et~al.]{liu2026agenticwam}
Z.~Liu, D.~Niu, B.~Zhang, B.~Shao, J.~Wang, R.~Zheng, Y.~Sharma, R.~Lu, J.~Liu,
  R.~Zhang, et~al.
\newblock Agentic world action models.
\newblock \emph{arXiv preprint arXiv:2610.30136}, 2026.

\bibitem[Luo et~al.(2026)Luo, Chen, Dimakis, Grauman, Chu, and
  Tran]{luo2026predictive}
M.~Luo, Y.~Chen, A.~Dimakis, K.~Grauman, W.-S. Chu, and D.~Tran.
\newblock Predictive structure improves video diffusion dynamics.
\newblock In \emph{European Conference on Computer Vision}, pages 515--533.
  Springer, 2026.

\bibitem[Ma et~al.(2026)Ma, Liao, Wang, Wang, Feng, Hu, Bao, Xi, Gan, Wang,
  et~al.]{ma2026scaling}
S.~Ma, J.~Liao, X.~Wang, J.~Wang, C.~Feng, Z.~Hu, C.~Bao, Z.~Xi, Y.~Gan,
  W.~Wang, et~al.
\newblock Scaling mixture-of-experts video pretraining for embodied
  intelligence.
\newblock \emph{arXiv preprint arXiv:2607.07675}, 2026.

\bibitem[Margoni et~al.(2024)Margoni, Surian, and
  Baillargeon]{margoni2024violation}
F.~Margoni, L.~Surian, and R.~Baillargeon.
\newblock The violation-of-expectation paradigm: A conceptual overview.
\newblock \emph{Psychological Review}, 131\penalty0 (3):\penalty0 716, 2024.

\bibitem[{MiniMax}(2026)]{minimax2026h3}
{MiniMax}.
\newblock {MiniMax H3}: An open model breaking the boundaries between tasks and
  modalities, July 2026.
\newblock URL \url{https://www.minimax.io/blog/minimax-h3}.

\bibitem[Motamed et~al.(2026)Motamed, Culp, Swersky, Jaini, and
  Geirhos]{physicsiq}
S.~Motamed, L.~Culp, K.~Swersky, P.~Jaini, and R.~Geirhos.
\newblock Do generative video models understand physical principles?
\newblock In \emph{2026 IEEE/CVF Winter Conference on Applications of Computer
  Vision (WACV)}, pages 948--958. IEEE, 2026.

\bibitem[{NVIDIA}(2026)]{nvidia2026gr00t2}
{NVIDIA}.
\newblock {GR00T} 2: Scaling robot foundation models for generalization,
  dexterity, and adaptation, Nov. 2026.
\newblock URL \url{https://research.nvidia.com/labs/gear/gr00t-2}.

\bibitem[{OpenAI}(2024)]{sora}
{OpenAI}.
\newblock Sora system card, Dec. 2024.
\newblock URL \url{https://openai.com/index/sora-system-card/}.

\bibitem[Riochet et~al.(2018)Riochet, Castro, Bernard, Lerer, Fergus, Izard,
  and Dupoux]{intphys}
R.~Riochet, M.~Y. Castro, M.~Bernard, A.~Lerer, R.~Fergus, V.~Izard, and
  E.~Dupoux.
\newblock Intphys: A framework and benchmark for visual intuitive physics
  reasoning.
\newblock \emph{arXiv preprint arXiv:1803.07616}, 2018.

\bibitem[Spelke et~al.(1992)Spelke, Breinlinger, Macomber, and
  Jacobson]{spelke1992origins}
E.~S. Spelke, K.~Breinlinger, J.~Macomber, and K.~Jacobson.
\newblock Origins of knowledge.
\newblock \emph{Psychological review}, 99\penalty0 (4):\penalty0 605, 1992.

\bibitem[Stahl and Feigenson(2015)]{stahl2015observing}
A.~E. Stahl and L.~Feigenson.
\newblock Observing the unexpected enhances infants’ learning and
  exploration.
\newblock \emph{Science}, 348\penalty0 (6230):\penalty0 91--94, 2015.

\bibitem[T{\'e}gl{\'a}s et~al.(2011)T{\'e}gl{\'a}s, Vul, Girotto, Gonzalez,
  Tenenbaum, and Bonatti]{teglas2011pure}
E.~T{\'e}gl{\'a}s, E.~Vul, V.~Girotto, M.~Gonzalez, J.~B. Tenenbaum, and L.~L.
  Bonatti.
\newblock Pure reasoning in 12-month-old infants as probabilistic inference.
\newblock \emph{Science}, 332\penalty0 (6033):\penalty0 1054--1059, 2011.

\bibitem[Thrush et~al.(2022)Thrush, Jiang, Bartolo, Singh, Williams, Kiela, and
  Ross]{winoground}
T.~Thrush, R.~Jiang, M.~Bartolo, A.~Singh, A.~Williams, D.~Kiela, and C.~Ross.
\newblock Winoground: Probing vision and language models for visio-linguistic
  compositionality.
\newblock In \emph{2022 IEEE/CVF Conference on Computer Vision and Pattern
  Recognition (CVPR)}, pages 5228--5238. IEEE, 2022.

\bibitem[Todorov et~al.(2012)Todorov, Erez, and Tassa]{todorov2012mujoco}
E.~Todorov, T.~Erez, and Y.~Tassa.
\newblock {MuJoCo}: A physics engine for model-based control.
\newblock In \emph{2012 {IEEE/RSJ} International Conference on Intelligent
  Robots and Systems}, pages 5026--5033. IEEE, 2012.

\bibitem[Tong et~al.(2024)Tong, Liu, Zhai, Ma, LeCun, and Xie]{mmvp}
S.~Tong, Z.~Liu, Y.~Zhai, Y.~Ma, Y.~LeCun, and S.~Xie.
\newblock Eyes wide shut? exploring the visual shortcomings of multimodal llms.
\newblock In \emph{2024 IEEE/CVF Conference on Computer Vision and Pattern
  Recognition (CVPR)}, pages 9568--9578. IEEE, 2024.

\bibitem[Ullman et~al.(2017)Ullman, Spelke, Battaglia, and
  Tenenbaum]{ullman2017mind}
T.~D. Ullman, E.~Spelke, P.~Battaglia, and J.~B. Tenenbaum.
\newblock Mind games: Game engines as an architecture for intuitive physics.
\newblock \emph{Trends in cognitive sciences}, 21\penalty0 (9):\penalty0
  649--665, 2017.

\bibitem[Von~Helmholtz(1867)]{helmholtz1867}
H.~Von~Helmholtz.
\newblock \emph{Handbuch der physiologischen Optik}, volume~9.
\newblock L. Voss, 1867.

\bibitem[Wan et~al.(2025)Wan, Wang, Ai, Wen, Mao, Xie, Chen, Yu, Zhao, Yang,
  et~al.]{wan2025}
T.~Wan, A.~Wang, B.~Ai, B.~Wen, C.~Mao, C.-W. Xie, D.~Chen, F.~Yu, H.~Zhao,
  J.~Yang, et~al.
\newblock Wan: Open and advanced large-scale video generative models.
\newblock \emph{arXiv preprint arXiv:2503.20314}, 2025.

\bibitem[Wang et~al.(2026{\natexlab{a}})Wang, Cai, Chen, Chi, Sun, Dai, Hung,
  Guo, Ren, Yao, et~al.]{wang2026codeasworlds}
H.~Wang, Y.~Cai, W.~Chen, J.~Chi, H.~Sun, Q.~Dai, Y.-H. Hung, X.~Guo, J.~Ren,
  R.~Yao, et~al.
\newblock Code as worlds: Agentic discovery of executable world representations
  for physical reasoning.
\newblock \emph{arXiv preprint arXiv:2608.27549}, 2026{\natexlab{a}}.

\bibitem[Wang et~al.(2026{\natexlab{b}})Wang, Wang, Lin, Ji, Wiedemer, Gao,
  Luo, Qian, Huang, Hong, et~al.]{wang2026very}
M.~Wang, R.~Wang, J.~Lin, R.~Ji, T.~Wiedemer, Q.~Gao, D.~Luo, Y.~Qian,
  L.~Huang, Z.~Hong, et~al.
\newblock A very big video reasoning suite.
\newblock \emph{arXiv preprint arXiv:2602.20159}, 2026{\natexlab{b}}.

\bibitem[Warstadt et~al.(2020)Warstadt, Parrish, Liu, Mohananey, Peng, Wang,
  and Bowman]{blimp}
A.~Warstadt, A.~Parrish, H.~Liu, A.~Mohananey, W.~Peng, S.-F. Wang, and S.~R.
  Bowman.
\newblock Blimp: The benchmark of linguistic minimal pairs for english.
\newblock \emph{Transactions of the Association for Computational Linguistics},
  8:\penalty0 377--392, 2020.

\bibitem[Wolpert et~al.(1995)Wolpert, Ghahramani, and
  Jordan]{wolpert1995internal}
D.~M. Wolpert, Z.~Ghahramani, and M.~I. Jordan.
\newblock An internal model for sensorimotor integration.
\newblock \emph{Science}, 269\penalty0 (5232):\penalty0 1880--1882, 1995.

\bibitem[Wu et~al.(2025)Wu, Zou, Li, Huang, Yang, Tan, Peng, Wu, Xiong, Jiang,
  et~al.]{wu2025hunyuanvideo}
B.~Wu, C.~Zou, C.~Li, D.~Huang, F.~Yang, H.~Tan, J.~Peng, J.~Wu, J.~Xiong,
  J.~Jiang, et~al.
\newblock Hunyuanvideo 1.5 technical report.
\newblock \emph{arXiv preprint arXiv:2511.18870}, 2025.

\bibitem[Wu et~al.(2017{\natexlab{a}})Wu, Lu, Kohli, Freeman, and
  Tenenbaum]{wu2017vda}
J.~Wu, E.~Lu, P.~Kohli, B.~Freeman, and J.~Tenenbaum.
\newblock Learning to see physics via visual de-animation.
\newblock \emph{Advances in neural information processing systems}, 30,
  2017{\natexlab{a}}.

\bibitem[Wu et~al.(2017{\natexlab{b}})Wu, Tenenbaum, and Kohli]{wu2017neural}
J.~Wu, J.~B. Tenenbaum, and P.~Kohli.
\newblock Neural scene de-rendering.
\newblock In \emph{2017 IEEE Conference on Computer Vision and Pattern
  Recognition (CVPR)}, pages 7035--7043. IEEE, 2017{\natexlab{b}}.

\bibitem[Yang et~al.(2025)Yang, Teng, Zheng, Ding, Huang, Xu, Yang, Hong,
  Zhang, Feng, et~al.]{yang2025cogvideox}
Z.~Yang, J.~Teng, W.~Zheng, M.~Ding, S.~Huang, J.~Xu, Y.~Yang, W.~Hong,
  X.~Zhang, G.~Feng, et~al.
\newblock Cogvideox: Text-to-video diffusion models with an expert transformer.
\newblock In \emph{International Conference on Learning Representations},
  volume 2025, pages 83048--83077, 2025.

\bibitem[Yi et~al.(2019)Yi, Gan, Li, Kohli, Wu, Torralba, and
  Tenenbaum]{clevrer}
K.~Yi, C.~Gan, Y.~Li, P.~Kohli, J.~Wu, A.~Torralba, and J.~B. Tenenbaum.
\newblock Clevrer: Collision events for video representation and reasoning.
\newblock \emph{arXiv preprint arXiv:1910.01442}, 2019.

\bibitem[Yuille and Kersten(2006)]{yuille2006vision}
A.~Yuille and D.~Kersten.
\newblock Vision as bayesian inference: analysis by synthesis?
\newblock \emph{Trends in cognitive sciences}, 10\penalty0 (7):\penalty0
  301--308, 2006.

\end{thebibliography}

\clearpage
\appendix
\suppressfloats
\section{Task Details}
\label{app:tasks}

FlipTest has 10 simulated tasks (\cref{tab:sim-tasks}) and 15 real-world tasks (\cref{tab:real-tasks}).
Together they form Level~1 of the hierarchy (300 minimal pairs, 600 initial frames), and the simulated scenes also supply the controls and sweeps of Levels~2 and~3.
Models see only the clean initial frame.
As \cref{fig:fig2-benchmark-overview} shows, the tasks cover six families of everyday physics:
\emph{statics and toppling} (box at a desk edge, object drop, propped stack, block stack, seesaws);
\emph{size and fit} (pipe, wall gap, funnel slide, containment by size, bowl stacking, shape matching);
\emph{friction and rolling} (block on an incline, chain over an edge, loop-the-loop);
\emph{collision and momentum} (domino cascades, billiards, Newton's cradle);
\emph{free fall and trajectory} (containment by distance, ring toss, water pouring, hand sanitizer);
and \emph{fluids and materials} (brush stroke, dye drop).

\begin{table}[t]
\centering
\caption{\textbf{Simulated tasks} (15 pairs each).
The two members of a pair sit at $\pm|\delta|$ around the threshold and differ only in the transition variable.
The 15 pairs of a task span the full $|\delta|$ range, from knife-edge to clear-cut cases.}
\label{tab:sim-tasks}
\fttable
\begin{tabularx}{\linewidth}{lXlll}
\toprule
\fthead \textbf{Task} & \textbf{Transition variable $\delta$} & \textbf{Threshold} & \textbf{$|\delta|$ range} & \textbf{Outcome labels} \\
\midrule
Box at desk edge & COM distance past the edge & $0$ & 0.008--0.045\,m & \texttt{stay} / \texttt{fall} \\
\ftzebra Ball through pipe & bore radius $-$ ball radius & $0$ & 0.0025--0.014\,m & \texttt{blocked} / \texttt{pass} \\
Ball through gap & slot width $-$ ball diameter & $0$ & 0.008--0.018\,m & \texttt{stuck} / \texttt{pass} \\
\ftzebra Domino & first-gap ratio $d/h$ & $1.0136$ & 0.10--0.50 & \texttt{full\_fall} / \texttt{break} \\
Chain over edge & hang fraction $f$ & $1/3$ & 0.03--0.13 & \texttt{rest} / \texttt{slide} \\
\ftzebra Loop-the-loop & release height $h$ & $0.776$\,m & 0.03--0.22\,m & \texttt{falls} / \texttt{completes} \\
Seesaw & torque share of one side & $1/2$ & 0.04--0.14 & \texttt{left\_down} / \texttt{right\_down} \\
\ftzebra Block on incline & tilt $\theta$ & $26.57^\circ$ & $1.5^\circ$--$8.0^\circ$ & \texttt{rest} / \texttt{slide} \\
Block stack & step $/$ block width & $1/3$ or $1/2$ & 0.20--0.32 & \texttt{stable} / \texttt{collapse} \\
\ftzebra Funnel slide & throat gap $-$ ball diameter & $0$ & 0.003--0.013\,m & \texttt{stuck} / \texttt{finished} \\
\bottomrule
\end{tabularx}
\end{table}

\begin{table}[t]
\centering
\caption{\textbf{Real-world tasks} (10 pairs, 20 frames each).}
\label{tab:real-tasks}
\fttable
\begin{tabularx}{\linewidth}{lXl}
\toprule
\fthead \textbf{Task} & \textbf{Transition variable} & \textbf{Outcome labels} \\
\midrule
Drop (desk edge) & overhang past the edge & \texttt{stay} / \texttt{fall} \\
\ftzebra Containment (size) & object size vs.\ aperture & \texttt{fits} / \texttt{does\_not\_fit} \\
Containment (distance) & release offset vs.\ cup & \texttt{in} / \texttt{out} \\
\ftzebra Domino & gap spacing in the row & \texttt{full\_fall} / \texttt{break} \\
Billiards & cue-ball aim vs.\ target & \texttt{hit} / \texttt{miss} \\
\ftzebra Newton's cradle & number of raised balls (1/2/3) & \texttt{1} / \texttt{2} / \texttt{3\_balls\_out} \\
Propped stack & stagger vs.\ support hull & \texttt{stays\_standing} / \texttt{collapses} \\
\ftzebra Brush stroke & pigment-loaded vs.\ clean brush & \texttt{colored\_stroke} / \texttt{no\_mark} \\
Water drop & pour offset vs.\ cup or glass & \texttt{fills} / \texttt{misses\_cup}; \texttt{into} / \texttt{misses\_pool} \\
\ftzebra Hand sanitizer & palm offset under the nozzle & \texttt{gel\_on\_palm} / \texttt{misses\_hand} \\
Ring toss & ring offset vs.\ post & \texttt{threads\_on} / \texttt{misses\_post} \\
\ftzebra Dye drop & dye vs.\ clear liquid & \texttt{colored\_plume} / \texttt{no\_visible\_change} \\
Bowl stack & which cup is held (size order) & \texttt{stacks} / \texttt{no\_stack} \\
\ftzebra Seesaw (real) & cube distances from the pivot & \texttt{left\_down} / \texttt{right\_down} / \texttt{balanced} \\
Matching game & object--cavity congruence & \texttt{fit} / \texttt{no\_fit} \\
\bottomrule
\end{tabularx}
\end{table}

\paragraph{Simulated tasks.}
We build all simulated scenes in MuJoCo and render them at $1280\times720$.
Within a task, the 15 magnitudes $|\delta|$ are evenly spaced over the range of \cref{tab:sim-tasks} and assigned to pairs in a fixed shuffled order.
In every pair, the first member lies at $-|\delta|$ and the second at $+|\delta|$, so each task contains both knife-edge and clear-cut cases.
To vary the viewpoint while keeping the two members aligned, we jitter the task's canonical camera in azimuth ($\pm18^\circ$), elevation ($\pm8^\circ$), and distance ($\times0.85$--$1.20$) and apply the \emph{same} jitter to both members of a pair.
A per-pair seed sets the visual theme (materials, colours, environment).
All thresholds are analytic except the domino threshold, which we calibrate by bisection in simulation.
The fixed scene constants include a 0.2\,m cube for the box, a bore radius of 0.06\,m for the pipe, a 0.12\,m slot for the gap, and a 0.075\,m throat for the funnel slide.
Friction is $\mu=0.5$ for the chain, so $f_c=\mu/(1{+}\mu)=1/3$, and for the incline, so $\theta_c=\arctan\mu$.
The loop radius is $R=0.30$\,m with ball radius $a=0.02$\,m, so $h_c=2.7R-1.7a$.
The domino row has 10 tiles, and all gaps after the first are spaced at $d/h=0.45$, so only the first gap crosses the boundary.
The stack task uses 3 blocks of width $w=0.10$\,m, and the direction of the two steps cycles over the pairs (staircase or zigzag), which changes the critical offset from $w/3$ to $w/2$.
In the funnel-slide task, the wall profile cycles over straight, arc, hourglass, and wave shapes, and the narrowest width always sets the boundary.
We simulate every member, and a pair enters the benchmark only if both simulated outcomes agree with the intended analytic side.
All 300 members pass this check.
The per-pair manifests record the exact scene parameters and signed margins, so every simulated example supports distance-to-boundary analysis.

\paragraph{Real-world tasks.}
The real split combines staged capture with controlled counterfactual edits.
Of the 300 real frames, 59 are initial frames of footage that we staged around the transition and filmed with a static phone camera.
The other 241 frames are counterfactual edits of captured or product photographs, or staged synthetic stills, and bowl stack, dye drop, matching game, and the real seesaw are built entirely this way.
In edited and synthetic pairs, the second member is an edit of the first that changes only the variable that decides the outcome (\eg the cup that is held in bowl stack, or the object in front of a fixed mold in the matching game).
The label is therefore fixed by construction.
Edits pass automated consistency checks and are regenerated on failure.
For the propped-stack scenes, we first set the stagger in MuJoCo, with a critical offset of $w/N=0.024$\,m and member offsets of 0 and 0.028\,m, and then restyle the renders photorealistically.
Newton's cradle and the real seesaw have three outcomes, and all other real tasks are binary.

\paragraph{Control and sweep probes.}
\Cref{fig:fig5-selectivity-controls}a and \Cref{fig:fig6-boundary-sweep}a show on the box task how Levels~2 and~3 extend a minimal pair into a graded probe.
A control group places four equally spaced states at margins $-3\delta, -\delta, +\delta, +3\delta$.
Only the middle step crosses the boundary, so the group separates flips caused by the boundary from flips caused by any small change.
A sweep places 30 states per scene across the boundary, densest near $m=0$, and traces where a model places the boundary and how sharply it switches.

\section{Outcome Judging}
\label{app:judge}

\paragraph{Blind protocol.}
The judge sees only the generated video and a text prompt.
It never sees the ground-truth label, the generation prompt, or the initial frame, so its reading depends on the video alone.
The prompt describes the scene, lists every outcome label of the task with a one-line description, and adds \texttt{unclear} (``not enough is shown to decide'').
It asks the judge which outcome the video \emph{actually depicts}:
\begin{quote}\small
\emph{You are watching a short generated video. The scene: \{scene\}. Decide which outcome the video ACTUALLY DEPICTS: \{label\} -- \{description\} \dots; ``unclear'' -- not enough is shown to decide. Judge only what is visibly shown, not what should physically happen. Reply as JSON with fields: outcome, confidence (0--1), reason (one sentence).}
\end{quote}
For simulated tasks, task rules make the labels precise and route broken physics to \texttt{unclear}.
For example, a box counts as \texttt{fall} as soon as it starts to descend over the edge, a pipe counts as \texttt{pass} only if the whole ball enters the top opening, and a domino row is \texttt{full\_fall} only if every original tile ends fallen.
Duplicated, merged, morphing, or teleporting objects, or a pivot that moves, give \texttt{unclear}.
The real tasks use the label descriptions alone (labels in \cref{tab:real-tasks}), and both judges use the same prompt.

\paragraph{Settings.}
Two judges, Gemini 3.8 Flash and Gemini 3.6 Flash, run at temperature 0 with a JSON response schema that restricts the outcome to the task labels plus \texttt{unclear}.
We sample videos at 8 frames per second, which captures sub-second events such as a ball dropping into a pipe that the default sampling of one frame per second misses.
Rate-limit and server errors are retried with exponential backoff.

\paragraph{Scoring rule.}
On pairs, a frame counts as correct only if both judges return the ground-truth outcome, and a pair is solved only if both of its frames are correct.
A frame on which the judges disagree, or that either judge rates \texttt{unclear}, counts as wrong, because a world model used for planning must commit to a clear outcome.
On controls and sweeps, the analyses average over many frames, so we use the labels of both judges (\cref{sec:metrics}).

\paragraph{Agreement between the judges.}
Over the $11\times600=6600$ videos of the main benchmark, the two judges return the same label for 85.8\% of videos (83.5\% on simulated tasks, 88.2\% on real tasks), with Cohen's $\kappa=0.85$ over the pooled labels.
Gemini 3.8 Flash is the stricter judge and rates 16.7\% of videos \texttt{unclear}, against 9.4\% for Gemini 3.6 Flash.
The dual rule counts a disagreement as wrong, so it can only lower accuracy relative to either judge alone, and every pair accuracy we report is conservative.
For Gemini Omni Flash 1.1, it lowers real pair accuracy from 34.7\% (Gemini 3.8 Flash alone) to 31.3\%.

\paragraph{Agreement with human labels.}
We manually labeled 600 generated videos sampled from the main benchmark with the same outcome labels plus \texttt{unclear}, and Gemini 3.6 Flash returns the human label for 79\% of them.
The main conclusions rest on gaps far larger than this readout noise.
Video models stay at or below 31.3\% on real pairs while the agents exceed 90\%.
The controls compare flip rates of the same model under the same judges, so a judge error that is independent of the side of the boundary affects causal and spurious flips alike.

\section{Agent Setup}
\label{app:agent}

\paragraph{Input.}
All three predictors (the direct VLM, the Measure agent, and the Simulation agent) are built on Gemini 4 Argon and receive exactly the same input.
The input is the clean initial frame, saved as an RGB PNG without metadata under an anonymous file name, together with the scenario text of the video prompt.
This text states the action and lists the outcomes without the camera instructions, and it names the allowed labels but never the decisive quantity.
Ground truth, sample identifiers, and source paths stay outside the agent's environment.
We run 780 jobs per predictor, covering the 300 real frames and the 480 sweep frames.

\paragraph{VLM, direct.}
Gemini 4 Argon answers in a single call without tools, at temperature 0 with a JSON schema (label, confidence, one-sentence reason).
It receives this prompt:
\emph{``This image is time zero, immediately before a physical event. Scenario: \{scenario\}. Predict the outcome that will ACTUALLY happen, using only this image.''}
All 780 calls returned a valid label.

\paragraph{Measure and Simulation agents (coding agent).}
The Measure agent and the Simulation agent use the same agent, Gemini 4 Argon at medium effort, and differ only in the method that their prompts prescribe.
They run in a command-line coding harness with a shell, file tools, an image viewer, and a Python environment (NumPy, SciPy, scikit-image, OpenCV, PIL).
Both prompts state that the image shows time zero and that the agent must work only from this image, without searching for the benchmark, other images, labels, or external implementations.
\begin{itemize}
\item \textbf{The Measure agent} writes a deterministic measurement script, \texttt{measure.py}, reports the key quantities that it measured, and decides the outcome with its own physical reasoning.
Physics engines and 3D simulation libraries are not available and must not be used.
\item \textbf{The Simulation agent} completes five steps.
(1) It measures the image with code.
(2) Before it builds any model, it writes a pre-simulation judgement \texttt{pre\_sim.json} from the measurements alone and never changes it.
(3) It builds the scene from scratch in MuJoCo with a deterministic \texttt{build\_scene.py} and states an assumption for every parameter that it cannot read from the image.
(4) It runs the simulation with \texttt{simulate.py}, which reads the outcome from the simulated state in code.
(5) It reports that outcome.
The agent may declare scenes that MuJoCo cannot represent (liquids, ink, gels, deformables) not simulable and answer them by reasoning.
\end{itemize}

\paragraph{Sandbox and audit.}
Each run executes in an isolated sandbox with no network except the model API.
The agent's shell cannot reach any other host, API keys are removed from its environment, the input directory is read-only, and only the output directory is writable.
For the Measure agent, an empty directory covers the installed physics-engine packages such as MuJoCo, so the agent cannot import them.
The harness also provides a server-side web-search tool that the network sandbox cannot block, so we audit every saved trace.
Any call to a web tool, or any network code (HTTP clients, sockets, package installation, API hosts) in text or files written by the agent, makes the run invalid.
No run used a web tool or failed the audit.

\paragraph{Budget.}
The main turn has a time limit of 40 minutes for the Measure agent and 45 minutes for the Simulation agent, and on timeout the agent gets one final five-minute turn to write its best answer.
This timeout happened in 1 run of the Measure agent and 4 runs of the Simulation agent.
We retry infrastructure failures once (5 runs of the Measure agent, 1 of the Simulation agent), and all 780 runs of each agent completed.
The median run takes 615\,s (Measure agent) or 636\,s (Simulation agent), uses about 115k tokens (about 28k output), and makes 23--24 tool calls.

\paragraph{Re-executing the Simulation agent.}
We score the Simulation agent on an independent re-execution of its code.
After each run, we copy \texttt{build\_scene.py} and \texttt{simulate.py} into a fresh offline sandbox that contains only the input image, rebuild the scene, run the simulation with a 15-minute limit, and count calls to \texttt{mj\_step}.
The score uses the outcome of this re-execution.
A simulated run is valid only if the re-execution succeeds, returns an allowed label, takes at least 200 simulation steps, and has at least one dynamic degree of freedom.
All 761 simulated runs passed.
Every rebuilt scene was byte-identical to the submitted scene, and every re-executed outcome matched the agent's answer (median 1500 steps).
File times and the trace confirm that the agent wrote \texttt{pre\_sim.json} before the first simulation step.

\paragraph{Scoring.}
Invalid runs (no answer, malformed answer, failed re-execution, or failed audit) count as wrong.
In the seesaw scenes, the prompt refers to the red and blue blocks, and we map their labels to the camera-relative \texttt{left\_down} and \texttt{right\_down} before scoring.

\subsection{Agent Analysis}
\label{app:agent-analysis}

\begin{figure}[tp]
    \centering
    \includegraphics[width=\textwidth]{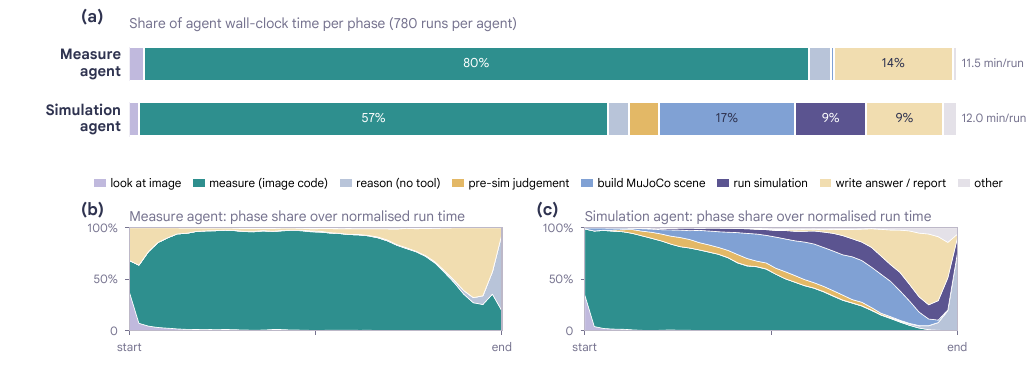}
    \caption{\textbf{Where the coding agents spend their time.}
(a) Share of wall-clock time per phase, pooled over the 780 runs of each agent; we assign each model step and its tool call to one phase from the call content.
(b, c) Phase share over normalised run time (start to end of each run, averaged over runs).
Both agents spend most of their time measuring the image, and the Simulation agent spends about a quarter of its run building and running the MuJoCo scene.}
    \label{fig:figA9-agent-analysis}
\end{figure}

\begin{figure}[tp]
    \centering
    \includegraphics[width=\textwidth]{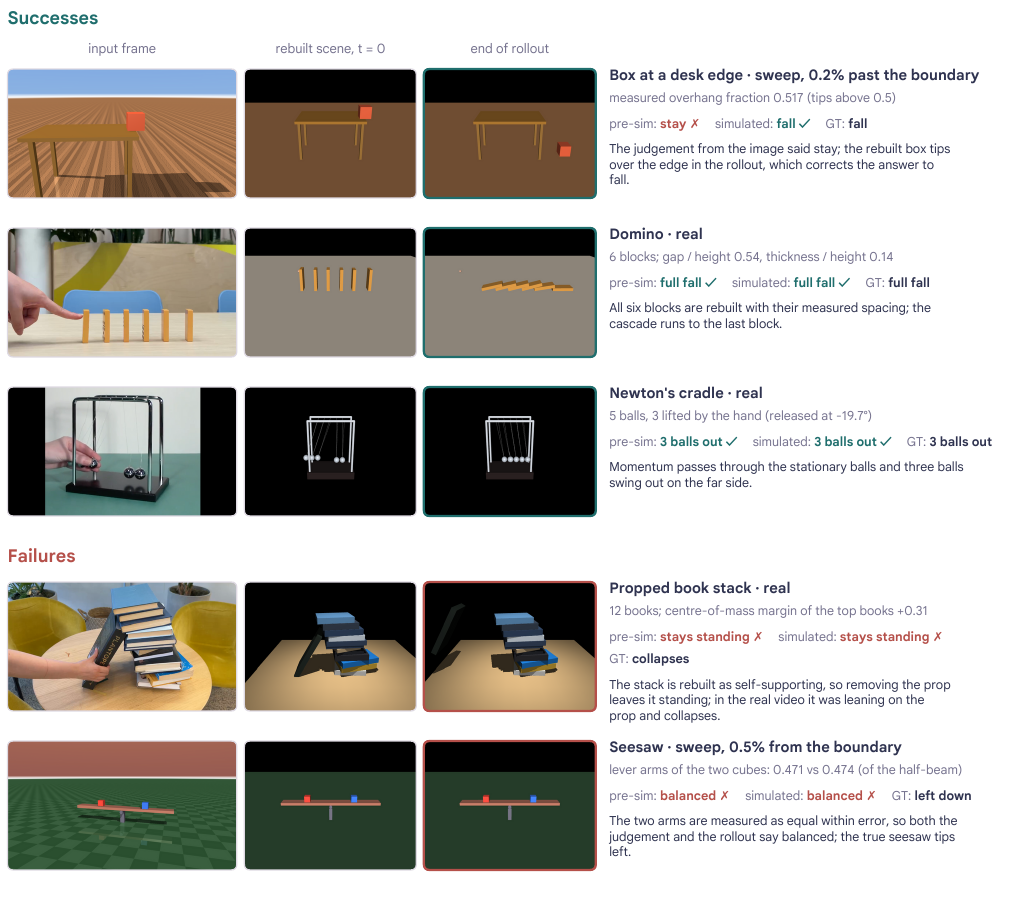}
    \caption{\textbf{Simulation agent case studies.}
For each run, we show the input frame and the agent's rebuilt MuJoCo scene at $t=0$ and at the end of the rollout, re-rendered offline from the submitted \texttt{scene.xml}, together with the key measurement, the pre-simulation judgement, the simulated outcome, and the ground truth.
Top: the rollout corrects a wrong judgement (box) or confirms a correct one in multi-body scenes (domino, Newton's cradle).
Bottom: the errors come from the reconstruction, where the scene misses a support (propped book stack) or reads two lever arms closer than the measurement error as equal (seesaw).}
    \label{fig:figA10-sim-agent-cases}
\end{figure}

\paragraph{Measurement removes the direct VLM's one-sided errors.}
The errors of the direct VLM are one-sided within a task.
It calls 42 resting blocks on the incline sliding and 10 staying boxes falling, and on the block stack it calls 25 collapsing towers stable.
With code-based measurement, the same model makes none of these errors.
The Measure agent is perfect on the box, incline, and stack sweeps, and its only three sweep errors are seesaw frames at the two nearest steps ($|\text{torque share}-\tfrac12|\le0.007$), where it answers \texttt{balanced}.
Its sweep accuracy is 94\% at the nearest step, 97\% at the next, and 100\% from the third step on.
On real pairs, its largest gains over the direct VLM occur when the decisive quantity is a length or an offset read from the image, as for brush strokes (50\%$\to$100\%), objects at the table edge (70\%$\to$100\%), and the domino row (90\%$\to$100\%).
It helps least when the decisive variable is three-dimensional and hard to measure from one view, and propped stacks (60\%) and the shape-matching game (70\%) remain its weakest tasks.
The Simulation agent errs in the same places.
Its nine sweep errors are seesaw frames at the two nearest steps, where both the pre-simulation judgement and the rebuilt scene give \texttt{balanced}, and on real pairs it is weakest on propped stacks (50\%) and the matching game (70\%).

\paragraph{Measurement dominates the run time.}
A run takes 11.5 minutes on average for the Measure agent and 12.0 minutes for the Simulation agent, with about 24 tool calls (\cref{app:agent}).
\Cref{fig:figA9-agent-analysis} splits this time into phases.
The Measure agent spends 80\% of its time writing and running image-measurement code and 14\% writing its answer and report, while reasoning without a tool call (3\%) and viewing the image (2\%) take little time.
The Simulation agent spends 57\% measuring, 17\% building the MuJoCo scene, 9\% running the simulation, 4\% writing the pre-simulation judgement, and 9\% writing its answer, and it moves through these phases in the order its prompt prescribes.
Turning pixels into numbers is therefore the main cost for both agents, and the physics rollout adds only about a tenth of the run time.

\paragraph{The Measure agent revises toward the correct answer.}
The prompt asks the Measure agent to write a first estimate to \texttt{answer.json} early and to overwrite it as its measurements improve.
In 56 of the 780 runs, the final outcome differs from the first one, and 52 of these revisions turn a wrong first estimate into a correct answer, against 4 in the other direction.
Measuring the image therefore corrects the visual first guess far more often than it overturns a correct one.

\paragraph{The rollout rarely changes the Simulation agent's answer.}
The Simulation agent simulated 761 of the 780 scenes and declared the other 19 not simulable (liquids, gels, and deformable objects), answering 18 of these 19 correctly by reasoning.
Its pre-simulation judgement, written from the measurements alone, is already correct on 96.0\% of real frames and 97.5\% of sweep frames.
The rollout overturned it in only 4 of the 761 simulated runs, with three corrections on sweep frames among the six closest to the boundary (box, incline, stack) and one error on a real bowl-stacking frame.
\Cref{fig:figA10-sim-agent-cases} shows typical cases.
Once the scene is right, the rollouts reproduce multi-body interactions faithfully (domino cascade, Newton's cradle).
The failures arise in the reconstruction, such as a stack rebuilt as self-supporting when it leans on a prop, or seesaw lever arms that differ by less than the measurement error.
As in \cref{sec:failure}, reading the physical state from the frame is the decisive step.

\section{Prompts}
\label{app:prompts}

\Cref{fig:figA5-prompt-video,fig:figA6-prompt-vlm,fig:figA7-prompt-agent,fig:figA8-prompt-judge} give the exact prompts for the box task (and the object-drop task for real prompts).
All other tasks use the same templates with their own scene description and labels.
The video world models receive the neutral generation prompt of \cref{fig:figA5-prompt-video}, which names both outcomes without stating the decision rule.
The vision-language models answer the yes/no question of \cref{fig:figA6-prompt-vlm} on the sweeps, and Gemini 4 Argon answers the direct prompt of \cref{fig:figA6-prompt-vlm} as the direct VLM of the agent study.
The coding agent receives the prompts of \cref{fig:figA7-prompt-agent}, and the judges receive those of \cref{fig:figA8-prompt-judge}.
Every model sees only the clean initial frame (or, for the judges, only the generated video), and annotated frames, ground truth, and margins are never shown.

\begin{figure}[t]
\centering
\promptbox{0.485}{Simulated task (box at desk edge)}{A rigid box rests on the edge of a table. Continue the video with real rigid-body physics: the box either stays put on the tabletop or tips over the edge and falls to the floor, depending on how far it overhangs. The camera is static: no zoom, no dolly, no pan. No new objects appear in the scene.}\hfill
\promptbox{0.485}{Real task (object drop)}{Nothing touches the object resting at the table edge -- no hand enters the frame and the table is never bumped. Show what happens to it next under gravity alone: either it stays resting on the table edge, or it tips over the edge and falls to the floor. The camera is static: no zoom, no dolly, no pan. No new object appears in the scene, no person or extra hand enters the frame, and nothing is nudged, steadied, straightened or repositioned to influence the outcome.}
\caption{\textbf{Prompts for the video world models.} Every model receives the clean initial frame and the text prompt of its task.
The prompt states the action and names both outcomes neutrally, without the decision rule or the side of the boundary.
Both members of a pair use the same prompt.}
\label{fig:figA5-prompt-video}
\end{figure}

\begin{figure}[t]
\centering
\promptbox{0.33}{Boundary sweep (Gemini 4 Argon, Gemini 3.8 Flash)}{The box on the table edge is left alone. Will it fall off the table? Answer with a single word: yes or no.}\hfill
\promptbox{0.635}{VLM, direct (Gemini 4 Argon)}{This image is time zero, immediately before a physical event.\\
\mbox{}\\
Scenario: A rigid box rests on the edge of a table. Under real rigid-body physics, the box either stays put on the tabletop or tips over the edge and falls to the floor, depending on how far it overhangs.\\
\mbox{}\\
Predict the outcome that will ACTUALLY happen, using only this image.\\
ALLOWED\_LABELS: fall, stay\\
\mbox{}\\
Reply as JSON with fields: label, confidence (0-1), reason (one sentence).}
\caption{\textbf{Prompts for the vision-language models.} Left: the yes/no question used for the boundary sweeps (\cref{sec:boundary}), where the answer \emph{yes} maps to the event label (here \texttt{fall}).
Right: the direct VLM of the agent study (\cref{sec:failure}) answers in a single call, and a JSON schema restricts the label to the allowed labels.
Both are shown for the box task.}
\label{fig:figA6-prompt-vlm}
\end{figure}

\begin{figure}[p]
\centering
\promptbox{0.485}{Measure agent}{\# Physical outcome prediction from one image\\
\mbox{}\\
You receive exactly one RGB image, @input.png in the current workspace (also at /task/input.png); view it with the image tool. It is time zero, immediately before the event. Work only from this image. Do not search for the benchmark, other images, labels, datasets or external implementations; you have no internet access beyond the model API.\\
\mbox{}\\
Scenario: A rigid box rests on the edge of a table. Under real rigid-body physics, the box either stays put on the tabletop or tips over the edge and falls to the floor, depending on how far it overhangs.\\
\mbox{}\\
ALLOWED\_LABELS: fall, stay\\
\mbox{}\\
METHOD (measurement only): write Python code that measures the image (NumPy, SciPy, scikit-image, OpenCV, PIL), then decide the outcome with your own physical reasoning from those measurements. Physics engines, rigid-body simulators and 3D scene/simulation libraries are not available and must not be used. Deliver /out/measure.py, deterministic, runnable as\\
\hspace*{1.0em}/opt/conda/bin/python /out/measure.py /task/input.png /out/measurements.json which writes the numeric measurements you relied on.\\
\mbox{}\\
Write /out/answer.json as soon as you have a first estimate and overwrite it when you refine~it.~Schema:\\
\{\textquotedbl{}outcome\textquotedbl{}: one of the allowed labels,\\
\hspace*{1.0em}\textquotedbl{}confidence\textquotedbl{}: number in [0, 1],\\
\hspace*{1.0em}\textquotedbl{}method\textquotedbl{}: \textquotedbl{}measured\textquotedbl{} (outcome follows from your measurements) or \textquotedbl{}reasoned\textquotedbl{} (you could not measure what matters and reasoned from the image),\\
\hspace*{1.0em}\textquotedbl{}key\_quantities\textquotedbl{}: \{\textquotedbl{}\textless{}name\textgreater{}\textquotedbl{}: \{\textquotedbl{}value\textquotedbl{}: \textless{}number\textgreater{}, \textquotedbl{}definition\textquotedbl{}: \textquotedbl{}\textless{}what was measured and how it is normalised\textgreater{}\textquotedbl{}\}, ...\},\\
\hspace*{1.0em}\textquotedbl{}notes\textquotedbl{}: short text\}\\
Also write /out/report.md with your measurements and~uncertainties.}\hfill
\promptbox{0.485}{Simulation agent}{\# Physical outcome prediction from one image\\
\mbox{}\\
You receive exactly one RGB image, @input.png in the current workspace (also at /task/input.png); view it with the image tool. It is time zero, immediately before the event. Work only from this image. Do not search for the benchmark, other images, labels, datasets or external implementations; you have no internet access beyond the model API.\\
\mbox{}\\
Scenario: A rigid box rests on the edge of a table. Under real rigid-body physics, the box either stays put on the tabletop or tips over the edge and falls to the floor, depending on how far it overhangs.\\
\mbox{}\\
ALLOWED\_LABELS: fall, stay\\
\mbox{}\\
METHOD (3D simulation with MuJoCo). Follow these steps in order:\\
1. Measure the image with code (NumPy, SciPy, scikit-image, OpenCV, PIL).\\
2. BEFORE building or running any MuJoCo model, write /out/pre\_sim.json:\\
\hspace*{1.5em}\{\textquotedbl{}outcome\textquotedbl{}: \textless{}allowed label\textgreater{}, \textquotedbl{}confidence\textquotedbl{}: \textless{}0-1\textgreater{}\} -- your judgement from the measurements alone. Never modify this file afterwards.\\
3. Decide whether the scene can be represented in MuJoCo (rigid bodies, joints, contacts). If it cannot (e.g. liquids, ink, gel, strongly deformable materials), use method \textquotedbl{}not\_simulable\textquotedbl{}, reason directly and go to step 6.\\
4. Build the scene from scratch; for every parameter you cannot read from the image (mass, friction, ...) state your assumption. Deliver:\\
\hspace*{2.5em}/out/build\_scene.py  deterministic; `/opt/conda/bin/python /out/build\_scene.py /task/input.png /out/scene.xml` writes the MJCF scene\\
\hspace*{2.5em}/out/scene.xml       its output (byte-identical on rebuild)\\
\hspace*{2.5em}/out/simulate.py     `/opt/conda/bin/python /out/simulate.py /out/scene.xml /out/sim\_result.json` runs the rollout with mujoco.mj\_step and writes \{\textquotedbl{}outcome\textquotedbl{}: \textless{}allowed label\textgreater{}, \textquotedbl{}evidence\textquotedbl{}: \{...\}\}\\
\hspace*{1.5em}Neither script may render or need a GPU. The runner re-executes both offline (15 min limit) and scores the outcome of that re-execution.\\
5. Read the outcome from the simulated state in code (not a hard-coded rule).\\
6. Write /out/answer.json (overwrite it whenever you refine~it).~Schema:\\
\{\textquotedbl{}pre\_sim\_outcome\textquotedbl{}: the outcome in /out/pre\_sim.json,\\
\hspace*{1.0em}\textquotedbl{}outcome\textquotedbl{}: one of the allowed labels (must equal sim\_result.json's outcome when method is \textquotedbl{}simulated\textquotedbl{}),\\
\hspace*{1.0em}\textquotedbl{}confidence\textquotedbl{}: number in [0, 1],\\
\hspace*{1.0em}\textquotedbl{}method\textquotedbl{}: \textquotedbl{}simulated\textquotedbl{} | \textquotedbl{}not\_simulable\textquotedbl{} | \textquotedbl{}reasoned\textquotedbl{},\\
\hspace*{1.0em}\textquotedbl{}key\_quantities\textquotedbl{}: \{\textquotedbl{}\textless{}name\textgreater{}\textquotedbl{}: \{\textquotedbl{}value\textquotedbl{}: \textless{}number\textgreater{}, \textquotedbl{}definition\textquotedbl{}: \textquotedbl{}\textless{}what was measured and how it is normalised\textgreater{}\textquotedbl{}\}, ...\},\\
\hspace*{1.0em}\textquotedbl{}scene\_params\textquotedbl{}: \{...\},\\
\hspace*{1.0em}\textquotedbl{}assumptions\textquotedbl{}: [\textquotedbl{}friction = ...\textquotedbl{}, ...],\\
\hspace*{1.0em}\textquotedbl{}sim\_result\textquotedbl{}: short description of what the simulation~showed\}\\
Also write /out/report.md with your measurements and~uncertainties.}
\caption{\textbf{Prompts for the coding agent (Measure and Simulation agents).} Both agents receive the same header (top of each box) and differ only in the method section.
The scenario and allowed labels are exactly those of the direct VLM.
Both prompts are shown for the box task.}
\label{fig:figA7-prompt-agent}
\end{figure}

\begin{figure}[t]
\centering
\promptbox{0.55}{Simulated task (box at desk edge)}{You are watching a short generated video. The scene: a rigid box resting on the edge of a table.\\
\mbox{}\\
Decide which outcome the video ACTUALLY DEPICTS:\\
\hspace*{1.0em}\textquotedbl{}fall\textquotedbl{}  - the box begins descending off the table edge, even if it is still falling when the video ends\\
\hspace*{1.0em}\textquotedbl{}stay\textquotedbl{}  - the box remains supported on the tabletop without beginning to fall\\
\hspace*{1.0em}\textquotedbl{}unclear\textquotedbl{} - not enough is shown to decide\\
\mbox{}\\
For this box task, apply these rules in order:\\
1. Track the ORIGINAL box from the first frame. If it duplicates, morphs into another box, or a new box appears, choose \textquotedbl{}unclear\textquotedbl{}. Do not count a falling duplicate as the original falling. If the tabletop or its supporting structure bends, breaks, or collapses, also choose \textquotedbl{}unclear\textquotedbl{} regardless of the box's apparent final position. These artifacts take priority over \textquotedbl{}fall\textquotedbl{} and \textquotedbl{}stay\textquotedbl{}.\\
2. If the original box launches or flies upward through the air, choose \textquotedbl{}unclear\textquotedbl{} even if it later lands off the table. An ordinary downward fall after tipping does not count as flying.\\
3. If the original box visibly starts descending off the intact table edge, choose \textquotedbl{}fall\textquotedbl{} immediately. It does NOT need to reach the floor or finish falling: a box still suspended in midair at the end counts as \textquotedbl{}fall\textquotedbl{} once downward motion has begun. Do not choose \textquotedbl{}stay\textquotedbl{} merely because the video ends before landing.\\
4. Choose \textquotedbl{}stay\textquotedbl{} only if the original box remains supported on the intact tabletop and never begins to fall.\\
\mbox{}\\
Judge only what is visibly shown, not what should physically happen. Reply as JSON with fields: outcome, confidence (0-1), reason (one sentence).}\hfill
\promptbox{0.42}{Real task (object drop)}{You are watching a short generated video. The scene: an object resting on a table near its edge.\\
\mbox{}\\
Decide which outcome the video ACTUALLY DEPICTS:\\
\hspace*{1.0em}\textquotedbl{}fall\textquotedbl{}  - the object tips over the edge and falls off the table\\
\hspace*{1.0em}\textquotedbl{}stay\textquotedbl{}  - the object stays resting on the table\\
\hspace*{1.0em}\textquotedbl{}unclear\textquotedbl{} - not enough is shown to decide\\
\mbox{}\\
Judge only what is visibly shown, not what should physically happen. Reply as JSON with fields: outcome, confidence (0-1), reason (one sentence).}
\caption{\textbf{Prompts for the VLM judges.} Both judges (Gemini 3.6 Flash and Gemini 3.8 Flash) receive only the generated video, sampled at 8 frames per second, and this prompt, with no ground truth and no initial frame.
Simulated tasks add task rules that make the labels precise and send broken physics (duplicated, morphing, or flying objects) to \texttt{unclear}.
Real tasks use the label descriptions alone.}
\label{fig:figA8-prompt-judge}
\end{figure}

\section{Additional Results}
\label{app:results}
\begin{figure}[t]
    \centering
    \includegraphics[width=\textwidth]{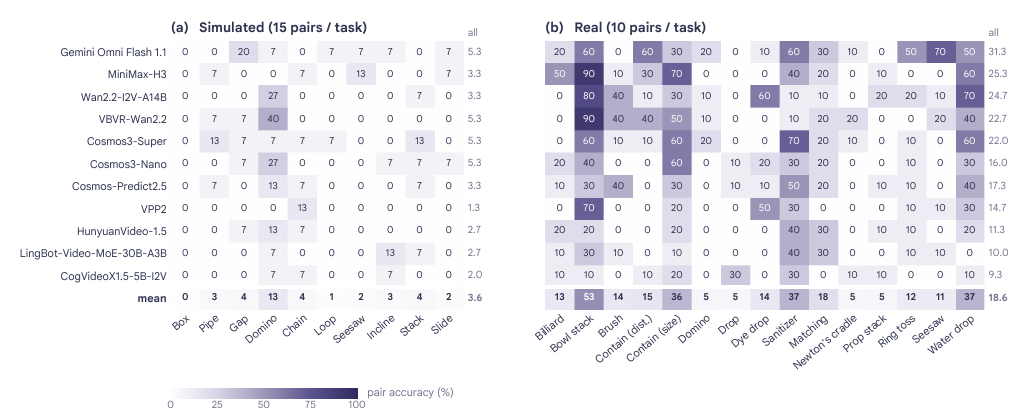}
    \caption{\textbf{Per-task pair accuracy (\%) of the 11 video models.}
(a) The 10 simulated tasks (15 pairs each); (b) the 15 real tasks (10 pairs each).
Rows are sorted as in \cref{fig:fig3-pair-accuracy}, the last row is the mean over models, and the \emph{all} column shows the split-level pair accuracy.
A frame counts as correct only if both judges (Gemini~3.8 Flash and Gemini~3.6 Flash) read the ground-truth outcome, as in \cref{fig:fig3-pair-accuracy}.}
    \label{fig:figA1-per-task-heatmap}
\end{figure}

\paragraph{Per-task breakdown.}
\Cref{fig:figA1-per-task-heatmap} shows that the failure holds across tasks.
On simulation, no task is solved reliably, the best task mean is 13.3\% (domino), and no model solves a single box pair.
Real tasks spread from 52.7\% (bowl stack) and 37.3\% (water drop, hand sanitizer) down to 4.5\% (drop, Newton's cradle).

\begin{figure}[t]
    \centering
    \includegraphics[width=\textwidth,height=0.86\textheight,keepaspectratio]{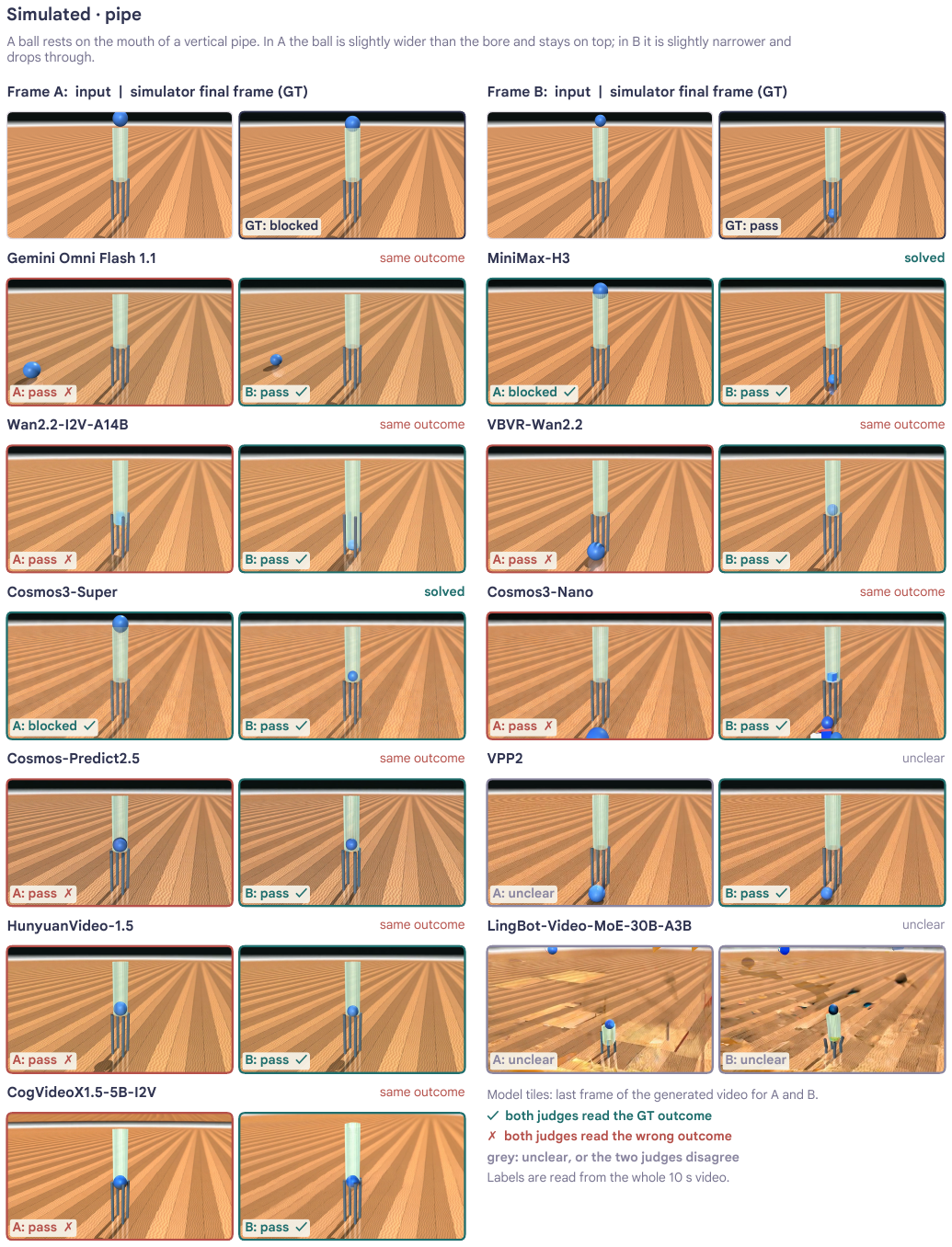}
    \caption{\textbf{All 11 video models on one simulated pipe pair.}
A ball rests on the mouth of a vertical pipe.
In A, the ball is slightly wider than the bore (GT: blocked), and in B it is slightly narrower (GT: pass).
The top row shows the input frames and the simulator's final frame, and each model tile is the last frame of that model's video.
A label shows the outcome read by the two judges, green when both read GT, red when both read the wrong outcome, and grey when the video is unclear or the judges disagree.
The tag above each model gives the pair result.}
    \label{fig:figA2-all-models-pipe}
\end{figure}

\begin{figure}[t]
    \centering
    \includegraphics[width=\textwidth,height=0.86\textheight,keepaspectratio]{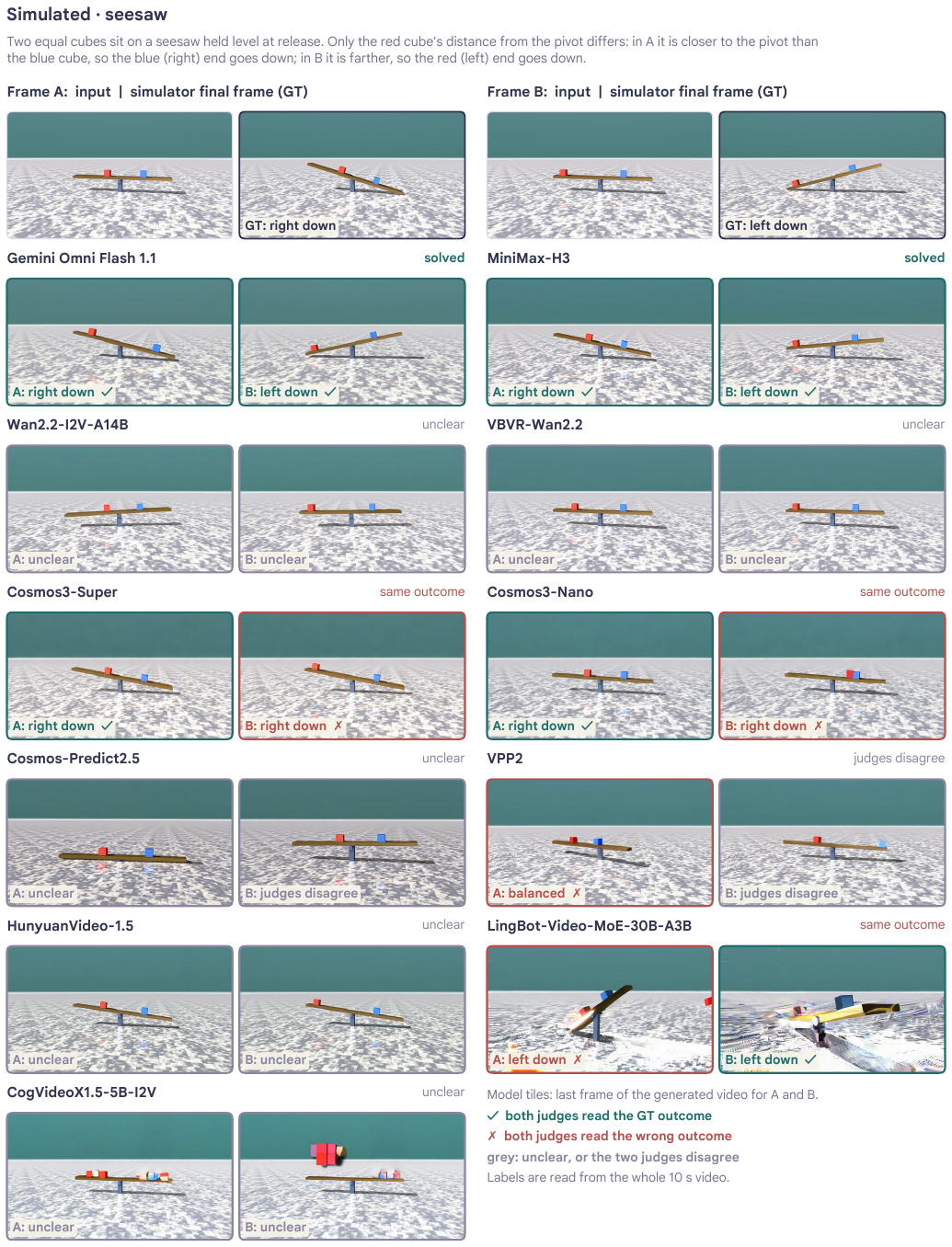}
    \caption{\textbf{All 11 video models on one simulated seesaw pair.} Two equal cubes sit on a seesaw that is level at release, and only the distance of the red cube from the pivot differs.
In A, the red cube is closer to the pivot than the blue cube (GT: right down), and in B it is farther away (GT: left down).
Layout as in \cref{fig:figA2-all-models-pipe}.}
    \label{fig:figA3-all-models-seesaw}
\end{figure}

\begin{figure}[t]
    \centering
    \includegraphics[width=\textwidth,height=0.86\textheight,keepaspectratio]{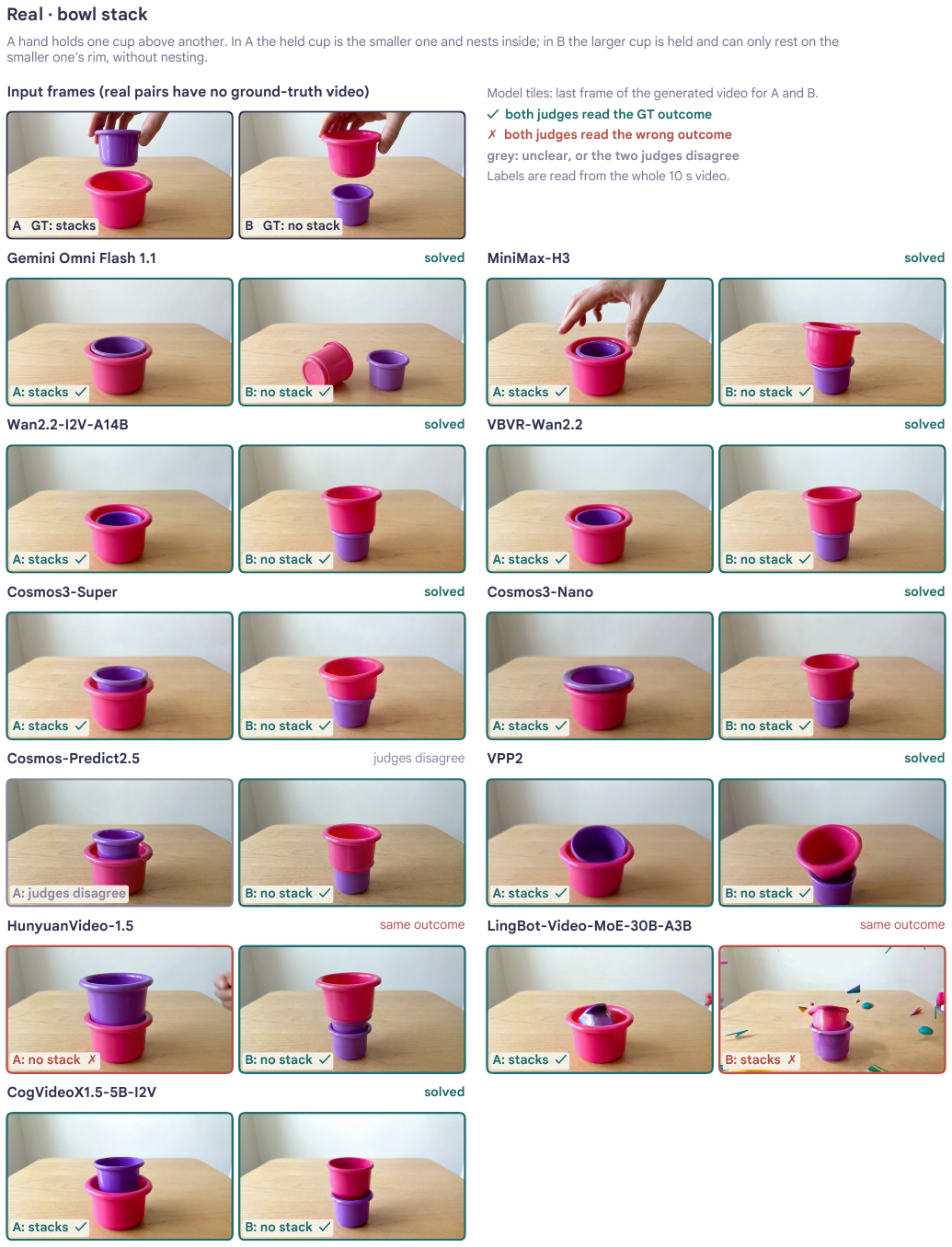}
    \caption{\textbf{All 11 video models on one real bowl-stack pair.} A hand holds one cup above another cup.
In A, the held cup is smaller and nests (GT: stacks), and in B it is larger and can only rest on the rim (GT: no stack).
Real pairs have no ground-truth video, so the top row shows only the inputs.
Layout as in \cref{fig:figA2-all-models-pipe}.}
    \label{fig:figA4-all-models-bowl-stack}
\end{figure}

\begin{figure}[p]
    \centering
    \includegraphics[width=\textwidth,height=0.86\textheight,keepaspectratio]{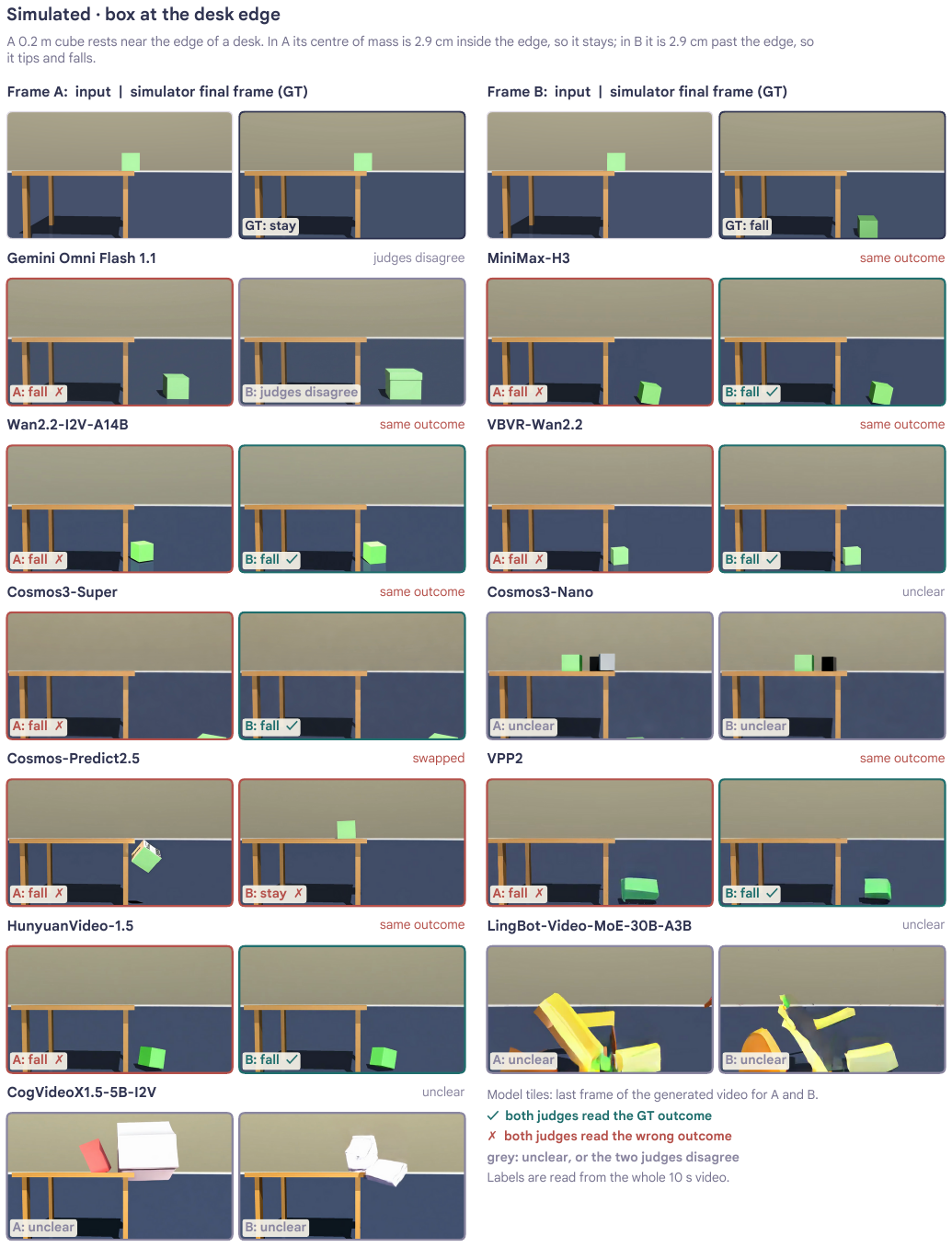}
    \caption{\textbf{All 11 video models on one simulated box pair.} A 0.2\,m cube rests near the edge of a desk, with its center of mass 2.9\,cm inside the edge in A (GT: stay) and 2.9\,cm past the edge in B (GT: fall).
No model solves the pair.
Layout as in \cref{fig:figA2-all-models-pipe}.}
    \label{fig:figA11-all-models-box}
\end{figure}

\begin{figure}[p]
    \centering
    \includegraphics[width=\textwidth,height=0.86\textheight,keepaspectratio]{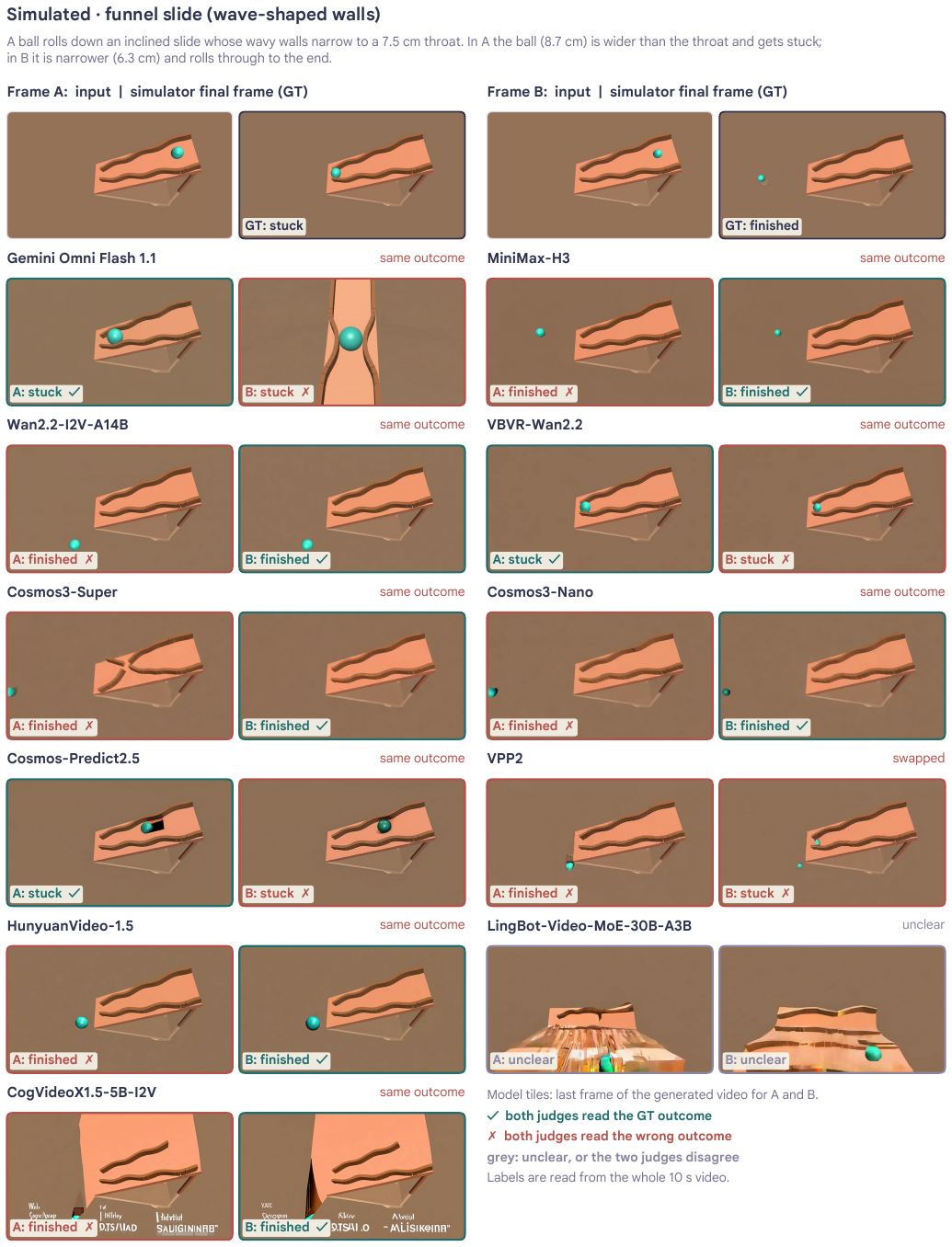}
    \caption{\textbf{All 11 video models on one simulated funnel-slide pair.} A ball rolls down a slide whose wave-shaped walls narrow to a 7.5\,cm throat.
In A, the ball (8.7\,cm) is wider than the throat (GT: stuck), and in B it is narrower (6.3\,cm; GT: finished).
Layout as in \cref{fig:figA2-all-models-pipe}.}
    \label{fig:figA12-all-models-slide}
\end{figure}

\begin{figure}[p]
    \centering
    \includegraphics[width=\textwidth,height=0.86\textheight,keepaspectratio]{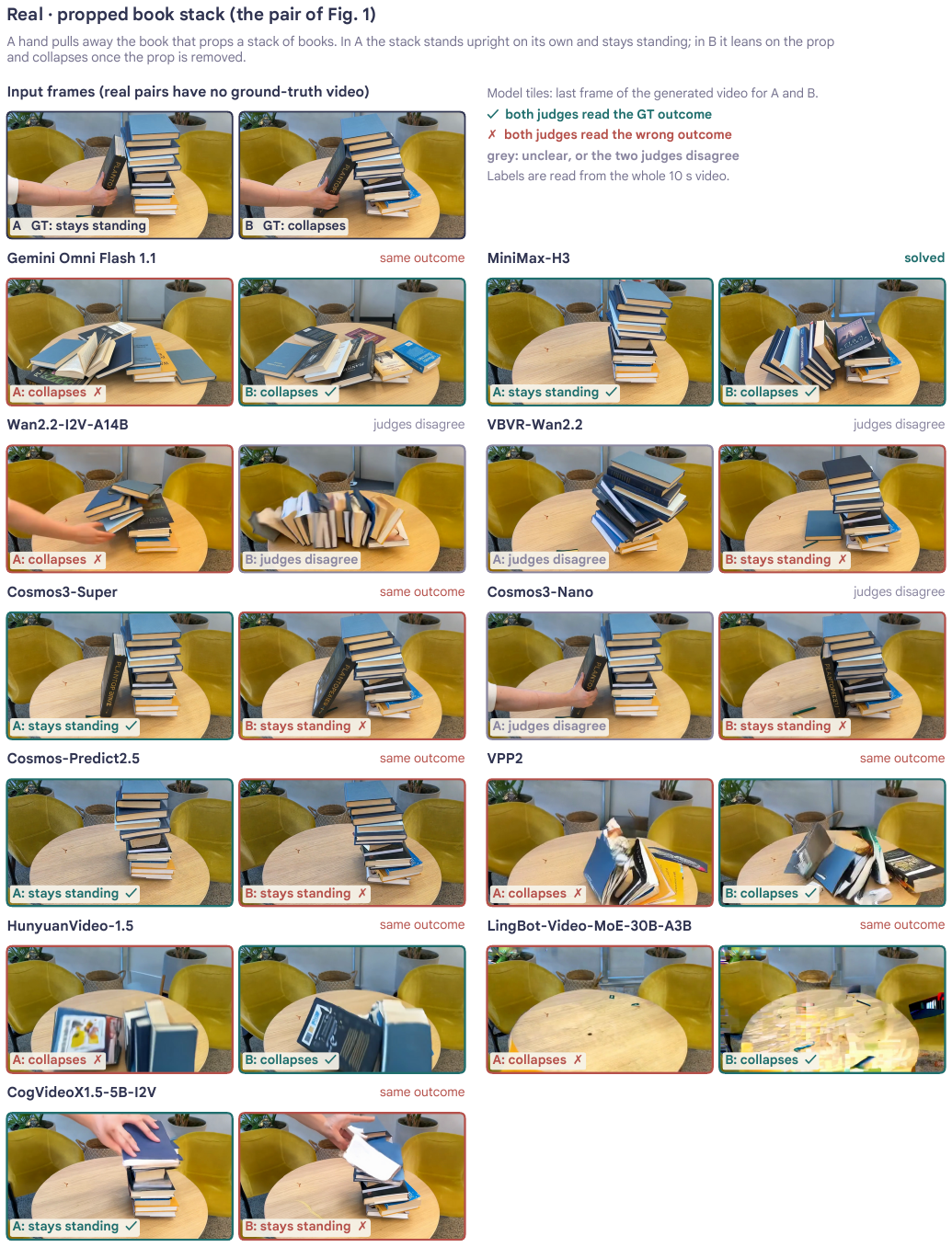}
    \caption{\textbf{All 11 video models on the real propped-stack pair of \cref{fig:fig1-teaser}.} A hand pulls away the book that props a stack.
In A, the stack stands upright on its own (GT: stays standing), and in B it leans on the prop (GT: collapses).
Real pairs have no ground-truth video, so the top row shows only the inputs.
Layout as in \cref{fig:figA2-all-models-pipe}.}
    \label{fig:figA13-all-models-prop-stack}
\end{figure}

\begin{figure}[p]
    \centering
    \includegraphics[width=\textwidth,height=0.86\textheight,keepaspectratio]{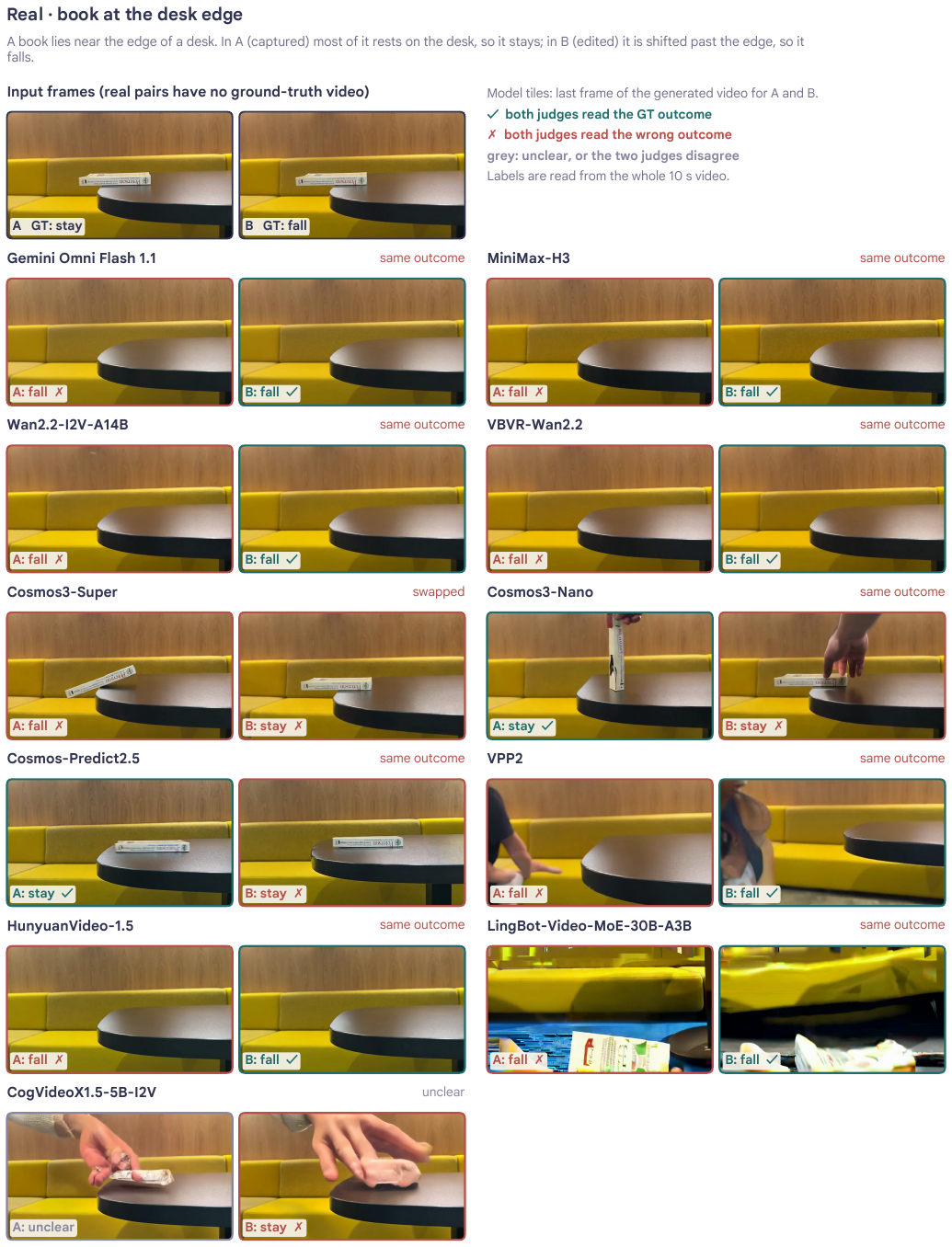}
    \caption{\textbf{All 11 video models on one real desk-edge pair.} A book lies near the edge of a desk.
In A (captured), most of the book rests on the desk (GT: stay), and in B (edited) the book shifts past the edge (GT: fall).
Layout as in \cref{fig:figA2-all-models-pipe}.}
    \label{fig:figA14-all-models-drop}
\end{figure}

\paragraph{All models on one pair.}
\Cref{fig:figA2-all-models-pipe,fig:figA3-all-models-seesaw,fig:figA4-all-models-bowl-stack,fig:figA11-all-models-box,fig:figA12-all-models-slide,fig:figA13-all-models-prop-stack,fig:figA14-all-models-drop} each show the predictions of all 11 video world models on a single minimal pair.
Each model tile is the last frame of the generated video, and the label is the outcome read from the whole video, scored as in \cref{fig:fig3-pair-accuracy}.
Most failures in these pairs show one outcome for both frames, as in \cref{fig:same-future}.
On the simulated pipe pair (\cref{fig:figA2-all-models-pipe}), 7 of 11 models let the ball drop through the pipe in both frames, only MiniMax-H3 and Cosmos3-Super solve it, and the judges find VPP2 and LingBot-Video-MoE-30B-A3B unclear.
On the simulated seesaw pair (\cref{fig:figA3-all-models-seesaw}), only Gemini Omni Flash 1.1 and MiniMax-H3 tip the beam to the correct side in both frames.
Cosmos3-Super and Cosmos3-Nano tip it to the right in both frames, and LingBot-Video-MoE-30B-A3B tips it to the left.
Five models keep the beam level or change the scene so much that the judges read \texttt{unclear} (Wan2.2-I2V-A14B, VBVR-Wan2.2, Cosmos-Predict2.5, HunyuanVideo-1.5, CogVideoX1.5-5B-I2V), and on VPP2 the judges disagree.
The real bowl-stack pair (\cref{fig:figA4-all-models-bowl-stack}) comes from the easiest real task in \cref{fig:figA1-per-task-heatmap}, and 8 of 11 models nest the small cup and leave the large cup resting on top.
The remaining models fail the same way, with LingBot-Video-MoE-30B-A3B stacking both cups and HunyuanVideo-1.5 stacking neither, and the judges disagree on Cosmos-Predict2.5.

The other four pairs show the same pattern.
No model solves the simulated box pair (\cref{fig:figA11-all-models-box}).
Six models make the box fall in both frames, Cosmos-Predict2.5 swaps the outcomes, the judges find three models unclear, and they disagree on Gemini Omni Flash 1.1.
On the simulated funnel slide with wave-shaped walls (\cref{fig:figA12-all-models-slide}), no model solves the pair either.
Nine models show the same outcome for both frames (three keep the ball stuck, and six let it roll through), VPP2 swaps the outcomes, and LingBot-Video-MoE-30B-A3B is unclear.
On the real propped book stack (\cref{fig:figA13-all-models-prop-stack}; also in \cref{fig:fig1-teaser}), only MiniMax-H3 solves the pair.
Seven models show the same outcome for both frames (four collapse the stack, three keep it standing), and the judges disagree on the other three.
On the real book at the desk edge (\cref{fig:figA14-all-models-drop}), nine models show the same outcome for both frames (seven drop the book, two keep it), Cosmos3-Super swaps the outcomes, and CogVideoX1.5-5B-I2V is unclear.

\paragraph{Outcome-conditioned generation.}
\Cref{fig:told-outcome} (\cref{sec:failure}) tests whether a video model renders the correct outcome once the prompt states it.
To keep the intervention minimal, each conditioned prompt replaces only the outcome clause of the original prompt, and in total 51 distinct conditioned prompts cover the 600 frames.
For example, the clause ``the box either stays put on the tabletop or tips over the edge and falls to the floor, depending on how far it overhangs'' becomes ``the box tips over the edge and falls to the floor'', and the rest of the prompt stays verbatim.
A single judge, Gemini 3.8 Flash, reads which outcome each video shows, and unclear videos count as wrong.

\paragraph{Per-model breakdown.}
\Cref{tab:per-model} splits the pairs of each model (150 simulated, 150 real) by outcome.
\emph{Same outcome} means that both judges agree on both frames, neither video is \texttt{unclear}, and the model shows the same outcome for both frames.
It is the most common result for every model except LingBot-Video-MoE-30B-A3B and CogVideoX1.5-5B-I2V on simulation, where unclear videos and judge disagreements dominate.
Swapped pairs (both opposite outcomes, on the wrong frames) are rare, at most 8 per model and split.
Models therefore seldom flip their prediction across a pair at all, and their main failure is insensitivity to the boundary.

\begin{table}[!htb]
\centering
\caption{\textbf{Outcome of every pair, per model and split.}
For \emph{solved}, both judges read the ground truth on both frames, and the pair accuracy of \cref{fig:fig3-pair-accuracy} is this count over 150.
For \emph{same}, both judges agree and the model shows the same outcome for both frames.
For \emph{swap.}, both opposite outcomes appear on the wrong frames, a category that also includes the few pairs with a different wrong outcome on a three-outcome task.
For \emph{uncl.}, both judges agree and at least one frame is \texttt{unclear}.
For \emph{disagr.}, the judges disagree on at least one frame.
Shaded columns give the solved counts, with the best count per split in bold.}
\label{tab:per-model}
\fttable\setlength{\tabcolsep}{3.5pt}
\begin{tabularx}{\linewidth}{>{\raggedright\arraybackslash}X >{\columncolor{ftteal!12}}r rrrr c >{\columncolor{ftteal!12}}r rrrr}
\toprule
\fthead & \multicolumn{5}{c}{\textbf{Simulated} (150 pairs)} && \multicolumn{5}{c}{\textbf{Real} (150 pairs)} \\
\cmidrule{2-6}\cmidrule{8-12}
\fthead \textbf{Model} & \textbf{solved} & same & swap. & uncl. & disagr. && \textbf{solved} & same & swap. & uncl. & disagr. \\
\midrule
Gemini Omni Flash 1.1 & \textbf{8} & 100 & 4 & 11 & 27 && \textbf{47} & 86 & 4 & 2 & 11 \\
MiniMax-H3 & 5 & 111 & 2 & 5 & 27 && 38 & 87 & 5 & 0 & 20 \\
Wan2.2-I2V-A14B & 5 & 85 & 5 & 17 & 38 && 37 & 82 & 8 & 1 & 22 \\
VBVR-Wan2.2 & \textbf{8} & 95 & 3 & 9 & 35 && 34 & 84 & 5 & 2 & 25 \\
Cosmos3-Super & \textbf{8} & 80 & 6 & 10 & 46 && 33 & 71 & 7 & 10 & 29 \\
Cosmos3-Nano & \textbf{8} & 65 & 6 & 25 & 46 && 24 & 77 & 8 & 4 & 37 \\
Cosmos-Predict2.5 & 5 & 74 & 5 & 16 & 50 && 26 & 76 & 2 & 7 & 39 \\
VPP2 & 2 & 91 & 2 & 14 & 41 && 22 & 94 & 3 & 1 & 30 \\
HunyuanVideo-1.5 & 4 & 89 & 4 & 22 & 31 && 17 & 91 & 6 & 2 & 34 \\
LingBot-Video-MoE-30B-A3B & 4 & 36 & 4 & 46 & 60 && 15 & 90 & 2 & 1 & 42 \\
CogVideoX1.5-5B-I2V & 3 & 31 & 1 & 62 & 53 && 14 & 74 & 5 & 8 & 49 \\
\bottomrule
\end{tabularx}
\end{table}

\section{A Possible Remedy: Contrastive Visual Schemata}
\label{app:schemata}

Our diagnosis points to a concrete remedy, which we leave to future work.
Video models often generate the same future for both frames of a pair (\cref{sec:pairs}), yet they render the correct outcome far more often once the prompt states it (\cref{sec:failure}).
They can draw the right event but fail to identify which condition in the initial frame decides it.
A \emph{contrastive visual schema} would point the model to that condition without giving away the answer.
It consists of two matched examples that differ in one causally relevant condition and lead to opposite outcomes, shown to the model alongside the target frame.
Before the model predicts a ball approaching an opening, for example, the schema would show a slightly smaller ball that passes through and a slightly larger ball that is blocked.
The examples come from a different scene, so they show which visual difference matters and leave the outcome of the target for the model to infer.
FlipTest already supplies a ready-made library of such schemata, since every minimal pair is a contrastive schema for its task.

The longer-term goal is a world model that finds the decisive condition without a hand-made schema for every new situation.
Minimal pairs could serve this goal as a training signal that keeps nearby states with different futures distinct in the model's representation, so that the model learns to locate the boundary on its own.

\section{Limitations}
\label{app:limitations}

FlipTest studies threshold dynamics of mostly rigid objects from a single initial frame.
We chose this setting because it makes ground-truth outcomes unambiguous and lets us locate the physical boundary precisely.
One decisive factor changes while the rest stays fixed, so any change in prediction can be attributed directly to that factor.
Real scenes often involve several interacting factors that unfold over time.
Current video models already miss boundaries set by a single factor, and extending FlipTest to scenes with several intersecting boundaries is a natural next step.

\end{document}